\documentclass{article}

\usepackage[utf8]{inputenc}
\usepackage{amsfonts}

\usepackage{algorithm}  % for algorithms
\usepackage{algorithmic} % for
\usepackage{float}
\usepackage{graphicx}

\usepackage[cmex10]{amsmath}
\DeclareMathOperator*{\argmax}{arg\,max}
\usepackage{amsthm}

\author{
   Paulo R. Cavalin\\
   \texttt{IBM Research}
   \and
   Marcelo N. Kapp\\
   \texttt{Universidade Federal da Integra\c{c}\~ao Latino Americana - Unila}
   \and
   Luiz S. Oliveira\\
   \texttt{Universidade Federal do Paran\'a - UFPR}
}
\title{Multi-scale Forest Species Recognition Systems for Reduced Cost}

\begin{document}

\maketitle

\begin{abstract}
This work focuses on cost reduction methods for forest species recognition 
systems. Current state-of-the-art shows that the accuracy of these systems have increased
considerably in the past years, but the cost in time to perform the recognition of
input samples has also increased proportionally. For this reason, in this work we focus
on investigating methods for cost reduction locally (at either feature extraction or 
classification level individually) and globally (at both levels combined), and evaluate
two main aspects: 1) the impact in cost reduction, given the proposed measures for it; 
and 2) the impact in recognition accuracy. The experimental evaluation conducted on
two forest species datasets demonstrated that, with global cost reduction, the cost of
the system can be reduced to less than 1/20 and recognition rates that are better than
those of the original system can be achieved.
\end{abstract}

\section{Introduction}
%The main purpose of this paper is to present and evaluate methods to reduce the costs of
%forest species recognition system, with minimum impact on the recognition accuracy reported
%in the state of the art.

Automatic forest species recognition has been drawing the attention of the research 
community, given its both commercial and environment-preserving value. For example, by
better monitoring wood timber trading, one may reduce commercialization of samples from species 
that are forbidden to be traded, e.g. species near extinction. However, in most cases, the wood
being traded has been cut into pieces of lumber, and identifying the forest species which
those wood lumbers come from generally requires an expert. For this reason, an automatic 
system for forest species recognition consists of an alternative to reduce the costs of 
hiring and training human experts and, hopefully, a way to improve the speed and accuracy in
performing this task.
%Recently various systems have been proposed 
%to cope with this problem 

In the image recognition community, we observe that forest species recognition has been 
generally treated as an image texture recognition problem 
\cite{ArunPriyaC2012,Bremananth2009,Gasson2010,Khalid2011,Martins2012b,Martins2013,PaulaFilho2010,Tou2009,Cavalin2013}. 
In contrast with image object recognition, where most of the shape of the object must 
be visible so as for a class to be associated to it, texture recognition can be conducted only on a small portion of
the whole. As a result, significant improvements in forest species recognition accuracy can be 
achieved with the use of multiple classifiers or multiple classifications. In \cite{PaulaFilho2010},
it is demonstrated that higher recognition rates can be achieved by dividing the images into sub-segments, 
and combining their individual classification results. Boost in the recognition rates can also be 
observed by making use of multiple feature sets \cite{Martins2012b}, or even by combining both ideas 
as in the multiple feature vector framework proposed in \cite{Cavalin2013}. 

Despite the improvement in recognition accuracy that multiple classifiers or multiple classifications can
bring to the forest species recognition task, a major drawback is the considerable increase in
the computational cost that is required to carry out the recognition of an input sample. When 
multiple feature sets are used, the cost increases linearly with the number of feature sets.
When the images are divided into sub-segments, the cost can increase quadratically. 

Given these standpoints, the main contribution of this paper lies in investigating and proposing methods to reduce 
the costs of this type of system using multiple classifications, with minimum impact on the 
recognition accuracy. To achieve this, we present investigations on cost reduction at 
different levels of a forest species recognition system. First, in Section~\ref{sec:classification_cost_reduction}, 
we introduce the Adaptive Multi-Level Framework (AMLF), which consists of an adaptive system 
for cost reduction at classification level. Next, in Section~\ref{sec:feature_extraction_cost_reduction},
we present an evaluation of cost reduction at feature extraction level, where the resolution
of the images are reduced to different scale ratios. Then, Section~\ref{sec:global_cost_reduction}
provides an analysis of cost reduction at global level, combining both feature extraction and 
classification cost reduction. It is worth mentioning that cost reduction for 
these evaluations are measured based on definition of cost presented in 
Section~\ref{sec:cost_definition}, where we focus on texture-based feature extraction methods 
and Support Vector Machine (SVM) classifiers. The results show that not only the 
cost can be very successfully reduced to about 1/20 of the original methods, but also better
recognition rates can observed with the proposed methods.

\section{State of the Art}
\label{sec:state_of_the_art}

\begin{figure}[H]
  \centering
  \includegraphics[width=5cm]{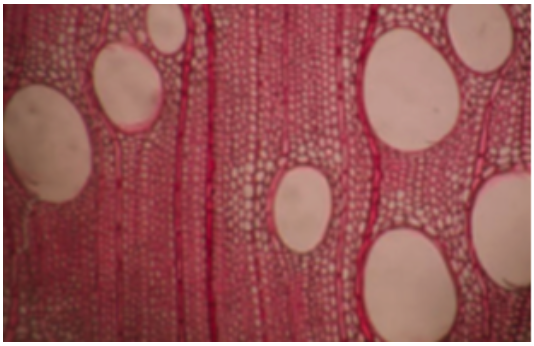}
  \caption{Samples of a microscopic image of Pinacae Pinus Taeda \cite{PaulaFilho2014}}
  \label{fig:samples1}
\end{figure}

\begin{figure}[H]
  \centering
  \includegraphics[width=5cm]{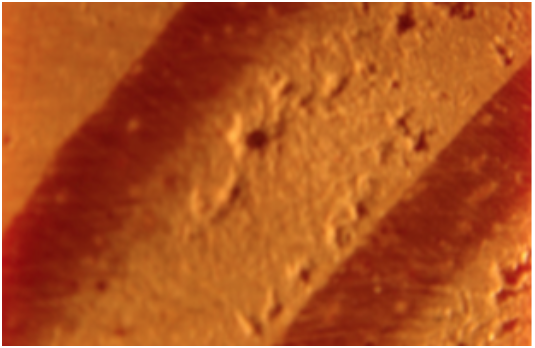}
  \caption{Samples of a macroscopic image of Pinacae Pinus Taeda \cite{PaulaFilho2014}}
  \label{fig:samples2}
\end{figure}

The recognition of forest species images can be divided into two approaches: microscopic 
and macroscopic. In the former, the image acquisition protocol is more complex since it 
depends on several procedures such as boiling the wood, cut it with a microtone, and 
dehydrating the slides, before acquiring the images. The result, though, is an image 
full of the details that can be useful to discriminate similar classes. This complex acquisition 
protocol, on the other hand, does not make the microscopic approach suitable to be used in the field, 
where one needs less expensive and more robust hardware \cite{PaulaFilho2014}. To overcome this 
problem, some authors have investigated the use of macroscopic images to classify forest species. 
Figure~\ref{fig:samples1} and Figure~\ref{fig:samples2} compares the microscopic and macroscopic samples of the the same forest 
species (Pinacae Pinus Taeda). As one may observe the macroscopic image has a significative loss 
of information when compared to the microscopic one.

Reviewing the literature of forest species recognition using macroscopic images, we may notice 
that the early works used mainly neural netwoks (multi-layer perceptron in most cases) and 
Gray-Level Co-ocurrence Matrices (GLCM) as features \cite{Tou07,Tou2008,Tou2009,Khalid08}. 
More recently, other representations such as Gabor filters \cite{yusof2010}, Local Binary 
Patterns (LBP) \cite{Nasirzadeh2010}, Local Phase Quantization (LPQ) \cite{PaulaFilho2014} and more 
robust classification schemes, such as ensembles of classifiers, were adopted, hence, raising 
the recognition rates. Systems based on representation learning using Convolutional Neural 
Networks (CNN) also have been exploited producing interesting results \cite{Hafemann2014}. 
Table \ref{comparisonI:tab} summarizes the works on forest species recognition using 
macroscopic images.

\begin{table} [h!]
\caption{Summary of the works on forest species recognition.}
\begin{center}
\begin{tabular}{llcc} \hline 
 \multicolumn{1}{c}{Authors.}&
 \multicolumn{1}{c}{Features} & 
 \multicolumn{1}{c}{Images/}&
 \multicolumn{1}{c}{Rec. Rate}\\ 
 \multicolumn{1}{c}{}&
 \multicolumn{1}{c}{}&
 \multicolumn{1}{c}{Classes}&
 \multicolumn{1}{c}{(\%)}\\ \hline
 
Tou et al.   \cite{Tou07}    			&   GLCM             &     360/5   & 72.0 \\ 
Tou et al.   \cite{Tou2008}   		 	&   GLCM, 1DGLCM     &  360/5      & 72.8 \\ 
Tou et al.    \cite{Tou2009}    			&   GLCM, Gabor, GLCM&          600/6   & 85.0 \\ 
Khalid et al. \cite{Khalid08} 			&   GLCM             &          1949/20 & 95.0 \\ 
Yusof et al.  \cite{yusof2010}  			&   Gabor, GLCM      &          3000/30 & 90.3 \\ 
Nasirzadeh et al. \cite{Nasirzadeh2010}   & LBP$^{u2}$, LBP$_{HF}$ &   3700/37 & 96.6 \\ 
Paula Filho et al. \cite{PaulaFilho2010}   &  Color, GLCM       &         1270/22     & 80.8 \\ 
Hafemann et al.    \cite{Hafemann2014}       & CNN                &         2942/41     & 95.7 \\ 
Paula Filho et al. \cite{PaulaFilho2014}   &  Color, CLBP, Gabor, LPQ&         2942/41     & 97.7 \\ \hline

\end{tabular}
\label{comparisonI:tab}
\end{center}
\end{table}

Research on microscopic imagens is more recent and became more popular with the release of a 
public database composed of 2240 images of 112 different species (see 
Section~\ref{sec:forest_species_datasets}), which made 
benchmarking and evaluation easier. Similarly to the literature on macroscopic images, most of the 
works on forest species using microscopic images use textural representation such as LBP 
\cite{Martins2012b}, LPQ \cite{Martins2012b} and their variants \cite{Kapp2014,Martins2015}. 
CNN also has been proved to be an interesting alternative for microscopic images \cite{Hafemann2014}. 
Table \ref{comparisonII:tab} summarizes the works on forest species recognition using microscopic images.

\begin{table} [htb]
\caption {Summary of the results published in the literature using the microscopic images of forest species}
\begin{center}
\begin{tabular}{llcc} \hline
 \multicolumn{1}{c}{Authors.}&
 \multicolumn{1}{c}{Features} & 
 \multicolumn{1}{c}{Images/}&
 \multicolumn{1}{c}{Rec. Rate}\\ 
 \multicolumn{1}{c}{}&
 \multicolumn{1}{c}{}&
 \multicolumn{1}{c}{Classes}&
 \multicolumn{1}{c}{(\%)}\\ \hline
 
Yadav et al. \cite{Yadav2013}      & Gabor+GLCM  					& 500/25  & 88.0-92.0 \\
Yusof et al. \cite{Yusof2013}      & Basic Gray Level Aura Matrix   & 5200/52 & 89.0-93.0  \\ 
Martins et al. \cite{Martins2012b} & LBP 							& 2240/112 & 80.7 \\
Martins et al. \cite{Martins2012b} & LPQ+LBP 						& 2240/112 & 86.5  \\
Cavalin et al. \cite{Cavalin2013}  & LPQ+GLCM 						& 2240/112 & 93.2 \\ 
Kapp et al. \cite{Kapp2014}        & LPQ+LPQ-Blackman+LPQ-Gauss   	& 2240/112 & 95.68 \\
Hafemann et al. \cite{Hafemann2014}& Convolutional Neural Network   & 2240/112 & 97.3 \\
Martings et al. \cite{Martins2015} & LBP+LPQ						& 2240/112 & 93.0 \\ 
Yadav et al. \cite{Yadav2015}      & Multiresolution LBP 			& 1500/75  & 97.4 \\ \hline
\end{tabular}
\label{comparisonII:tab}
\end{center}
\end{table}

\subsection{Datasets}
\label{sec:forest_species_datasets}
In this section we describe the two Forest Species databases used in this work, namely microscopic 
database and macroscopic database, 
respectively\footnote{The databases are freely available in 
http://web.inf.ufpr.br/vri/image-and-videos-databases/forest-species-database and 
http://web.inf.ufpr.br/vri/image-and-videos-databases/forest-species-database-macroscopic}. 

The microscopic database, which is described in detail in \cite{Martins2013}, contains 112 different 
forest species catalogued by the Laboratory of Wood Anatomy at the Federal University 
of Paran\'a in Curitiba (UFPR), Brazil. These images were acquired with an Olympus Cx40 
microscope equipped with a 100x zoom, after the wood went through some chemical/physical steps 
such as boiling, veneer coloring, and dehydration. In total, the dataset contains 2,240 
microscopic images, with a resolution of 1,024 by 768 pixels, equally distributed into set 
of 112 classes. It is worth mentioning that 37 of the classes correspond to Softwood species,
while 75 consist of Hardwood.

The macroscopic database was also catalogued by the Laboratory of Wood Anatomy at UFPR, and it is composed 
of 2,942 samples of 41 distinct species. In this case,
though, the images were captured with a Sony DSC T20 digital camera, resulting in image with a resolution
of 3,264 by 2,448. The number of samples per class ranges from 37 to 99, with an average of 71.75. 
Greater detail about this dataset can be found in \cite{PaulaFilho2014}. 

\section{General Cost Definition}
\label{sec:cost_definition}
The meaning of what we refer to as cost basically represents the time required for 
recognizing an image, or a set of test images (in considering that the different methods
involved in the cost analysis are evaluated on the same set). Although one can directly 
measure the running time of a given application to process such a set, and simply compare the 
differences among implementations of different systems, some factors such as operating system 
workload can affect this method even if the operating system, programming language and 
hardware used to implement and run 
the systems are exactly the same.

In consequence, we define cost in a more general form, which can be applied to any distinct
implementation, based on counting the total number of basic operations that are necessary 
to recognize the set of samples. The basic operation can be defined differently depending
on which part of the system we need to evaluate, or using an operation that is common for
all parts.

Let $Ne$ be the number of samples in the test set, in a general form, cost can be defined
as:
\begin{equation}
   \label{eq:cost}
   Cost = \sum_{i=1}^{Ne} ( Cost^F + Cost^C ),
\end{equation}
\noindent where $Cost^F$ corresponds to the cost of feature extraction, and $Cost^C$ to
the cost of classification.

In the next sections we present in details ways to measure $Cost^C$ and $Cost^F$,
respectively, and how reducing their values impacts recognition accuracy, either 
individually or combined.

\section{Classification Cost Reduction}
\label{sec:classification_cost_reduction}
In this section, we present the proposed approach named Adaptive Multi-Level Framework (AMLF), 
the main goal of which is to perform forest species recognition with accuracy close to the 
Single-Level Multiple Feature Vector Framework (SLF), proposed in \cite{Cavalin2013}, but with
reduced cost at classification level. 

Basically, AMLF consists of layers of different versions of SLF, with varied costs, and the main
idea is to rely on less costly\footnote{costly or expensive are terms that use interchangeably 
to express the same concept} but less accurate layers of SLF first, then move to more costly 
and consequently more accurate layers depending in the difficulty to recognize a given sample.

Before describing the aforementioned approaches, we first define a way to measure
classification cost.

\subsection{Classification Cost Definition}
\label{sec:classification_cost_def}
In this section we specify Equation~\ref{eq:cost} to compute cost at classification level,
allowing further to compare the cost of AMLF with different versions of SLF. Thus, given 
that feature extraction cost reduction is out of the scope of this section, we first adapt
that equation to consider only classification:
\begin{equation}
   Cost' = \sum_{i=1}^{Ne}  Cost^C.
\end{equation}

The previous equation is too general, so we need to define a way to compare the different 
approaches based on basic operations. Given that the main difference between the approaches
that we consider in this paper is the number of classifications performed for a sample, the
basic operation is herein defined as a classification step performed by a classifier, i.e. 
running a classification algorithm until a classification output comes out. Thus, let $f_i$ 
be a function that returns the number of classifications needed to recognize the $i$-th 
test sample, $Cost'$ can be defined as:
\begin{equation}
   Cost' = \sum_{i=1}^{Ne} f_i.
\end{equation}

Nonetheless, another cost factor must also be considered. Depending on the type of base classifier, a 
classification might be more costly than others, i.e. it might require more time to be conducted. 
For this reason, we extend the previous equation to:
\begin{equation}
   \label{eq:cost_classification}
   Cost' = \sum_{i=1}^{Ne}{ (f_i \times \Gamma_i )},
\end{equation}
\noindent where $\Gamma_i$ represent the cost of the base classifier, which may be, for instance, 
the size (the number of hidden neurons) of a Multilayer Perceptron Neural Network  or the number 
of support vectors for Support Vector Machines (SVMs) (similarly to the {\it total number of feature 
values} (TVF) measure \cite{Last2002}), to weigh the classification operation.

Next, we describe the framework presented in \cite{Cavalin2013} and how
Equation~\ref{eq:cost_classification} can be adapted for this specific system.

\subsection{Single-Level Multiple Feature Vector Framework (SLF)}
\label{sec:slf}
\begin{figure}[h]
   \centering
   \includegraphics[width=\textwidth]{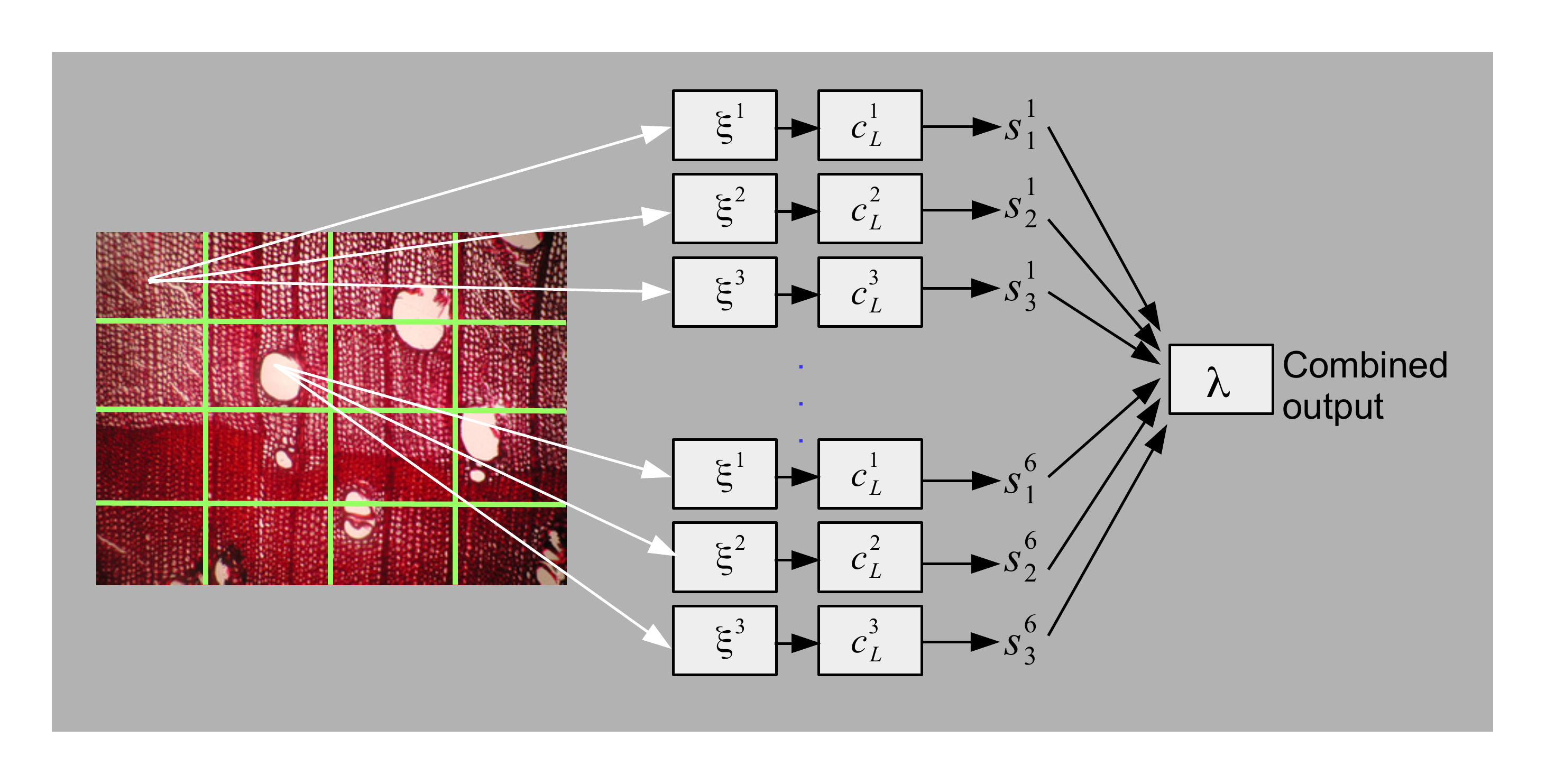}
   \caption{Overview of the SLF method. In this example, three feature sets are extracted 
   from $im'_1$ and $im'_6$, resulting in output $s^1_j$ and $s^6_j$.}
   \label{fig:slf_fig}
\end{figure}

The main idea of SLF lies in using information from multiple feature vectors
to improve recognition performance, owing to the variability introduced by these multiple vectors. 
One way to do so is with the extraction of diverse feature vectors by both dividing
the input image into sub-images (or patches, which is a term that we use interchangeably), and 
by combining different feature sets, as depicted in
Figure~\ref{fig:slf_fig}.

The main steps of SLF are listed in Algorithm~\ref{alg:framework}. The inputs
for this algorithm are: the original image, denoted $im$; the parameter $L$, to define 
the number of patches in which $im$ will be divided; the set of feature sets 
$\Xi= \{ \xi^1, \ldots, \xi^M\}$, where $\xi^j$ represents a distinct feature set with $m^j$ 
features, and $M = |\Xi|$ corresponds to the total number of feature sets; the set of classifiers 
$C_L = \{ c_L^1, \ldots, c_L^M\}$, corresponding to the set of classifiers trained for a 
given value of $L$ for each $\xi^j$; the number of classes of the problem denoted $K$; and 
a fusion function $\lambda$. From these inputs, the first steps consist in
dividing the input image into $N$ sub-images. That is, in step 2, the number of patches $N$ 
is computed as a function of $L$, e.g. $N = f(L)$. Then, in step 3, $im$ is divided into $N$ non-overlapping 
sub-images with identical sizes, generating the set $I = \{im'_1, \ldots, im'_N\}$. Next, the 
feature extraction is carried out. In steps 4 to 8, for each image $im'_i$ in $I$ and each feature 
set $\xi^j$ in $\Xi$, the feature vector $v_i^j$ is extracted and saved in $V$. Afterwards, 
each feature vector $v_i^j$ in $V$ is classified by the corresponding classifier $c_L^j$, resulting
in the scores $s_i^j(k)$ for every class $k$, where $1 \le k \le K$. 
Finally, all the scores $s_i^j(k)$ are combined using the combination function $\lambda$ and the 
final recognition decision $\phi$ is made, i.e. the forest species (a class $k$) from which $im$ 
has been extracted is outputted.

\begin{algorithm}[h]
\caption{The main steps of SLF.}
\label{alg:framework}
   \begin{algorithmic}[1]
      \STATE {\bf Input:} $im$, the input image; $L$, the parameter to compute the number of patches 
      to divide $im$; $\Xi = \{ \xi^1, \ldots, \xi^M\}$, the collection of feature sets, where $\xi^j$ 
      corresponds to a distinct feature set; $C_L = \{ c_L^1, \ldots, c_L^M\}$, the set of 
      classifiers trained for the given $L$, for each $\xi^j$; $K$, the number of classes of the problem;
     and $\lambda$, a fusion function to combine multiple classification results.
      \STATE $N = f(L)$
      \STATE Divide $im$ into $N$ non-overlapping sub-images with equal size, generating the set 
      $I = \{ im'_1, \ldots, im'_N\}$
      \FOR {each image $im'_i$ in $I$}
         \FOR {each feature set $\xi^j$ in $\Xi$}
            \STATE Extract feature vector $v_i^j$ from $im_i$ by considering $\xi^j$ as feature set, and 
            save $v_i^j$ in $V$
         \ENDFOR
      \ENDFOR
      \FOR {each feature vector $v_i^j$ in $V$}
         \STATE Recognize $v_i^j$ using classifier $c_L^j$, and save the scores $s_i^j(k)$ for 
     each class $k$, where $1 \le k \le K$.
      \ENDFOR
      \STATE Combine all scores $s_i^j(k)$ using $\lambda$, and compute
      output probabilities $P_L$.
   \end{algorithmic}
\end{algorithm}

We can adapt Equation~\ref{eq:cost_classification} in order to parametrize the cost function according
to the value of $L$, which affects the number of classifications, and the cost of the base classifier
trained for level $L$. Consider that $f(L)$ returns the number of classifier calls needed for recognizing a 
testing sample with SLF at level $L$, and $\Gamma(C_L)$ is a function that returns the cost 
of the classifiers in $C_L$. The cost $Cost'_{SFL}(L)$ for SLF with level set to $L$ can be 
calculated with Equation~\ref{eq:cost_slf}.
\begin{equation}
\label{eq:cost_slf}
   Cost'_{SLF}(L) =  Ne \times f(L)\times \Gamma(L)
\end{equation} 

As shown in \cite{Cavalin2013} (and also demonstrated in Section~\ref{sec:amlf_experiments}), the value
of $L$ can greatly affect both recognition performance and cost. While an increase of more than
10 percentage points can be observed in accuracy by changing $L$ from 1 to 3, the number of 
classifications increases from 1 to 16. The cost increase is quadratic in this case. Furthermore, 
the same cost is required for all test samples no matter the difficulty level to conduct the 
recognition of each image. These are the reasons that inspired us to propose the adaptive framework
described in Section~\ref{sec:amlf}.

\subsection{Adaptive Multi-Level Framework (AMLF)}
\label{sec:amlf}
\begin{figure}[h]
   \centering
   \includegraphics[width=\textwidth]{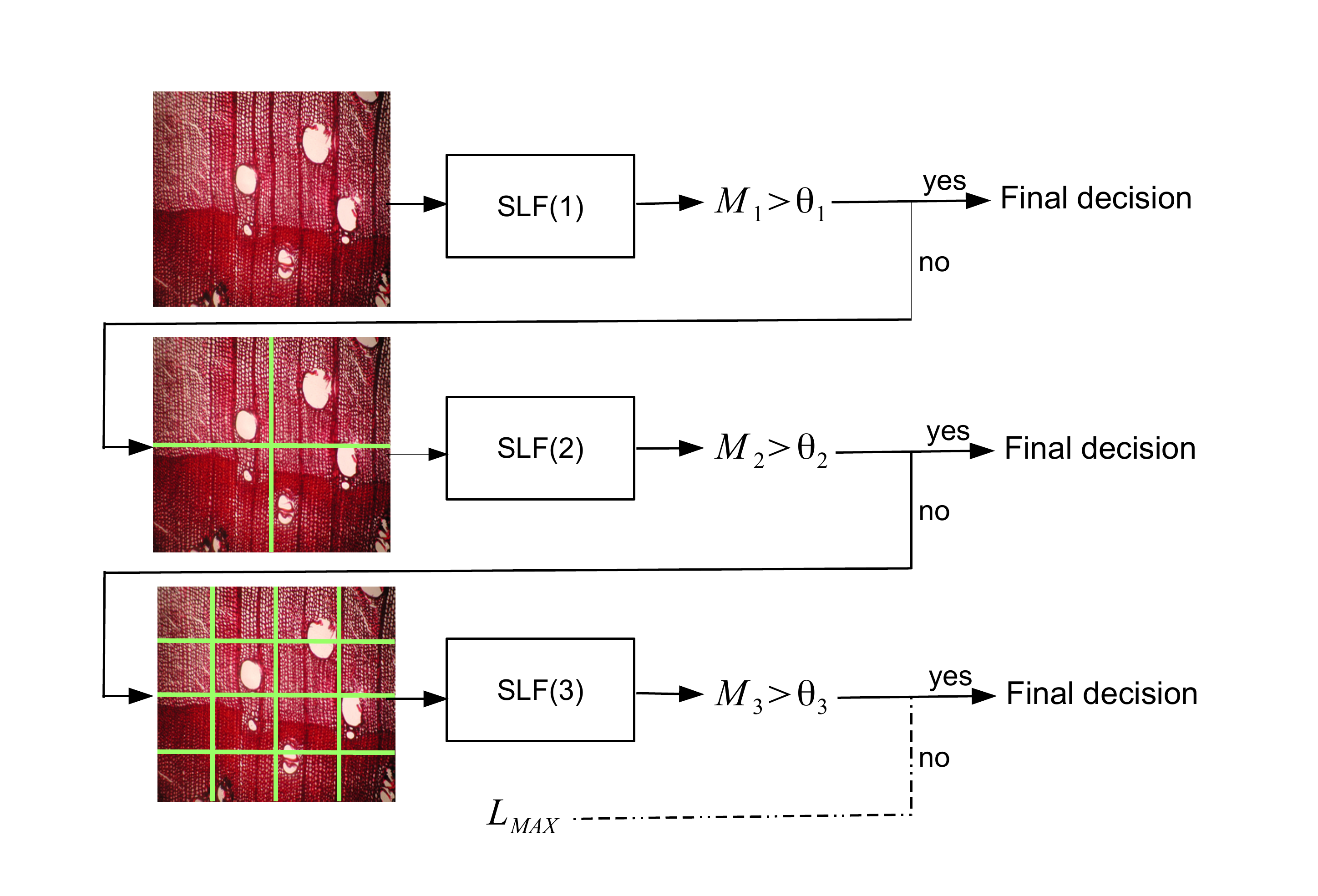}
   \caption{An illustration of AMLF.}
   \label{fig:amlf_fig}
\end{figure}

AMLF consists of evaluating consecutive layers of SLF, starting with layers with smaller values 
for $L$. If the level of confidence of the recognition result of such level is not high
enough, $L$ is incremented and layers with bigger values are evaluated until the maximum 
level $L_{max}$ is reached, as illustrated in Figure~\ref{fig:amlf_fig}. The level of 
confidence is computed based on the margin of the top two classes and a set 
of pre-defined thresholds, i.e. one threshold for each level $L < L_{max}$. 

This approach is better described in Algorithm~\ref{alg:multilevel_framework}. It 
consists of iterating the level value $l$ from 1 until $L_{max}$, the maximum value for $l$ (step 2). In step 3, 
Algorithm~\ref{alg:framework} (SLF) is called with the $L$ parameter set to $l$. The recognition
probabilities computed in this iteration for image $im$ are saved in $P_{l}$. Next, in step 4, the 
margin $M_{l}$ is computed from $P_{l}$. If the value in $M_{l}$ is above the pre-defined threshold
$\theta_{l}$ or the maximum level was reached, i.e. $l = L_{max}$, then the final recognition decision
$\phi$ is computed and the algorithm stops (steps 5 to 8).

\begin{algorithm}[h]
\caption{The AMLF approach.}
\label{alg:multilevel_framework}
   \begin{algorithmic}[1]
      \STATE {\bf Input:} $im$, the input image; $L_{max}$, the maximum number of layers; 
      $\Xi = \{ \xi^1, \ldots, \xi^M\}$, the collection of feature sets, where $\xi^j$ 
      corresponds to a distinct feature set; $C_L = \{ c_L^1, \ldots, c_L^M\}$, a set of 
      classifiers trained for each $1 \le l \le L_{max}$, for each $\xi^j$; $K$, the number of classes of the problem       
      $\lambda$, a fusion function to combine multiple classification results; $\{\theta_1, \ldots, \theta_{L_{max}-1}\}$,
      the rejection thresholds for each level.
      \FOR {$l$ from 1 to $L_{max}$}
            \STATE Call SLF (Algorithm~\ref{alg:framework}) with $L = l$ and save output probabilities in $P_{l}$
            \STATE Compute the margin $M_{l}$ using Equation~\ref{eq:margin}
            \IF {$M_{l} < \theta_{l}$ {\bf or} $l == L_{max}$}
                 \STATE Compute the final recognition decision $\phi$
                 \STATE {\bf Stop algorithm}
            \ENDIF
      \ENDFOR
   \end{algorithmic}
\end{algorithm}

Given that $P_{l} = \{p_1, \ldots, p_K\}$ represents the set of probabilities computed for image $im$ at level $l$, the 
margin $M_{l}$ is computed according to Equation~\ref{eq:margin}:
\begin{equation}
\label{eq:margin}
	M_{l} = p_{i} - p_{j}, 
\end{equation} 
where $ i = \argmax_{k=1}^{K} p_{k}$ and $ j = \argmax_{j=1}^{K} p_j \  \forall j \ne i$.

The rejection thresholds are defined on the validation set using a two-step procedure. The first step consists of 
computing the minimum and maximum values of margin observed in the validation set for each level $l$,
denoted $M'_{l}$ and $M''_{l}$, respectively. This process relies on SLF only. In the second step, we evaluate the
best combination of thresholds in ranges between $M'_{l}$ and $M''_{l}$, using grid search. In this case, AMLF is 
used.  After this process, the set of thresholds $\theta_l$ ($1 \le l < L_{max}$) that achieved the best 
recognition rates on the validation set are selected to
be used during operation.

The cost for an implementation of AMLF can be computed by extending Equation~\ref{eq:cost_slf}. 
Let $Ne_{L}$ be the number of samples
which AMLF exited at level $L$, and consider that the cost of recognizing theses samples correspond to
the same cost of $Cost'_{SLF}(L)$ in Equation~\ref{eq:cost_slf} where $Ne = Ne_{L}$ plus the cost of 
all previous levels, i.e. $\sum{Cost'_{SLF}(l)}$ for all $l < L$. The cost of AMLF can be computed as:
\begin{equation}
\label{eq:cost_mlf}
   Cost'_{AMLF}(L_{max}) = \sum_{L=1}^{L_{max}} [ Ne_{L} \times \sum_{l=1}^{L} ( f(l) \times \Gamma(l) ) ].
\end{equation} 

%\section{Methodology}
%\label{sec:methodology}
%In this section we first  describe how SLF can be extended to a multi-level structure, where multiple layers
%can be evaluated consecutively until the final decision is made. We also describe a methodology 
%to compare the complexity of this multi-level structure against SLF.

\subsection{Experiments}
\label{sec:amlf_experiments}
In this section, we present the experiments to evaluate both the recognition rates achieved
with AMLF, and the resulting cost associated to the system. These results are compared with
the different versions of SLF that were used to compose AMLF. Details on how the methods were
implemented can be found in \cite{Cavalin2015}.

\subsubsection{Protocol}
\label{sec:experiments_protocol}
We consider both microscopic and macroscopic databases described in 
Section~\ref{sec:forest_species_datasets}, and same experimental protocol defined in 
\cite{Cavalin2013}, allowing us to perform a direct comparison of the results.

The samples of each class have been partitioned in: 
50\% for training; 20\% for validation; and  30\% for test. Each subset has been randomly sampled with 
no overlapping between the sets. 
For avoiding the results to be biased to a given partitioning, this scheme is repeated 10 times. As a consequence, 
the results presented further represent the average recognition rate over 10 replications (each replication is related to different a partitioning).

As the base classifier, we make use of Support Vector Machines (SVMs) with Gaussian kernel\footnote{
in this work we used the LibSVM tool available at http://www.csie.ntu.edu.tw/{\tt \~{}}cjlin/libsvm/.}. 
Parameters $C$ and $\gamma$ were optimized by means of a grid search with hold-out validation, 
using the training set to train SVM parameters and the validation set to evaluate the performance. 
After finding the best values for $C$ and $\gamma$, an SVM was trained with both the training and
validation sets together. Note that normalization was performed by linearly scaling each attribute 
to the range [-1,+1].  

%For the LPQ feature set, we used the implementation provided by University of 
%Oulu, Department of Computer Science and Engineering, available athttp://www.cse.oulu.fi/CMV/Downloads/LPQMatlab.} 

%The experiments presented in the following paragraphs consider various implementations for 
%the proposed Multiple Feature Vector framework, whereas in each implementation we consider
%a different configuration for the group of feature sets.
%\footnote{We used the implementation provided by University of Oulu, Department of Computer
%Science and Engineering, available athttp://www.cse.oulu.fi/CMV/Downloads/LPQMatlab.} 
%In this case, for each implementation we show the impact of three different values for the parameter 
%$L$, i.e. 1, 2, and 3. Regarding the combination functions  described in Section~\ref{sec:combination_functions}, .

The comparison between SLF and AMLF takes into account twelve different systems. For SLF there are
9 different systems varying in terms of the parameter $L$, which is set to 1, 2 and 3, and in terms of feature
sets, given that we consider LBP and LPQ feature sets both individually and combined. For AMLF, we consider
three different implementations. We set $L_{max}$ to 3 and varied the feature set, also making use of LBP and LPQ
individually and combined. 

The thresholds $\theta_l$, $1 \le l < L_{max}$ were set with a grid search of AMLF applying on the validation
set, with classifiers trained only on the training set only to prevent from over fitting. 
%The thresholds found by this method are 
%listed in Table~\ref{tab:thres}.

%\begin{table}
%\centering
%\caption{The average of the thresholds found for AMLF with $L_{max} = 3$, on the validation set. Between parenthesis is
%the standard deviation. }
%\label{tab:thres}
%\begin{tabular}{|c|c|c|}
%   \hline
%   {\bf Feature set} & $\theta_1$ & $\theta_2$ \\
%   \hline
%   LBP & 0.074 (0.017) & 0.729 (0.117) \\
%   LPQ & 0.098 (0.015) & 0.783 (0.123) \\
%   LBP+LPQ & 0.449 (0.095) & 0.990 (0.029) \\
%   \hline
%\end{tabular}
%\end{table}

\subsubsection{Results on the Microscopic dataset}
\begin{figure}[H]
   \centering
   \includegraphics[width=9cm]{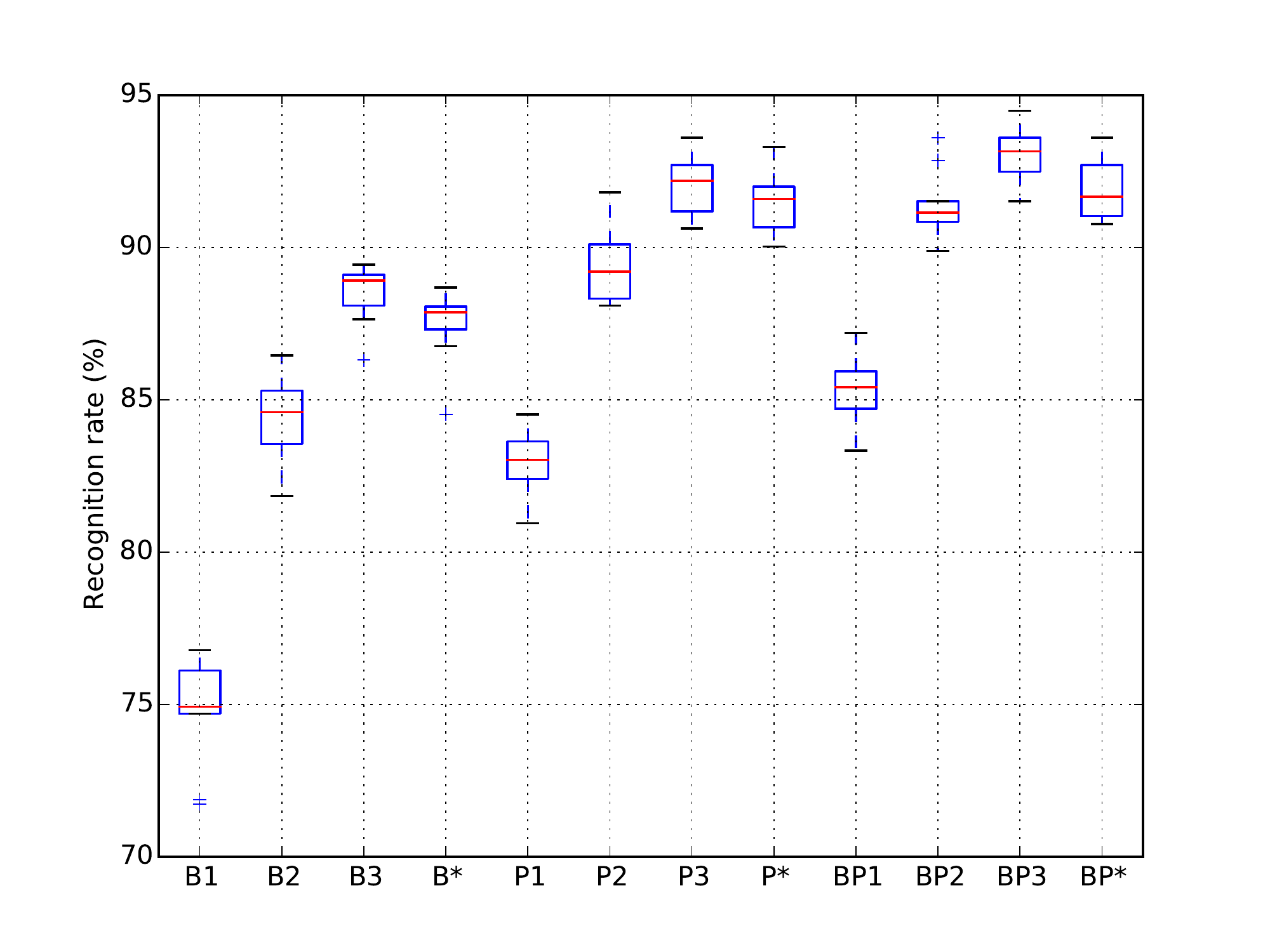}
   \caption{Recognition rates on the test set of the microscopic database, over 10 replications. 
   B1, B2, B3: SLF with LBP and $L$ set to 1, 2 and 
   3, respectively; B*: AMLF with LBP feature set; the same notation is used for P1, P2, P3, P*, for the LPQ feature
   set, and BP1, BP2, BP3, BP* for both LBP and LPQ combined.}
   \label{fig:rec_rates_micro}
\end{figure}

In Figure~\ref{fig:rec_rates_micro} the average recognition rates of the twelve aforementioned systems are presented.
The average recognition rates of 93.08\% were the best ones in these experiments, achieved by SLF with LBP and 
LPQ feature sets and $L$ set to 3. With LBP only, the accuracy presented by SLF were of 88.50\% with the
same value for $L$, and with LPQ, 92.03\%. The recognition rates of AMLF, with the same feature sets,
were of 91.90\%, 87,50\%, 91.44\%, for LBP and LPQ combined, LBP only, and LPQ only, respectively.
Despite a small loss in average accuracy of AMLF compared with SLF with $L=3$, the standard deviation of
the approaches show to that these systems resulted in similar recognition rates and, no significant 
loss of performance is observed with the use of AMLF. 
%Apparently, even
%though there is no loss of recognition performance, there is no gain either. Thus, one may question the
%advantage of such complex multi-level structure.

\begin{figure}[H]
   \centering
   \includegraphics[width=9cm]{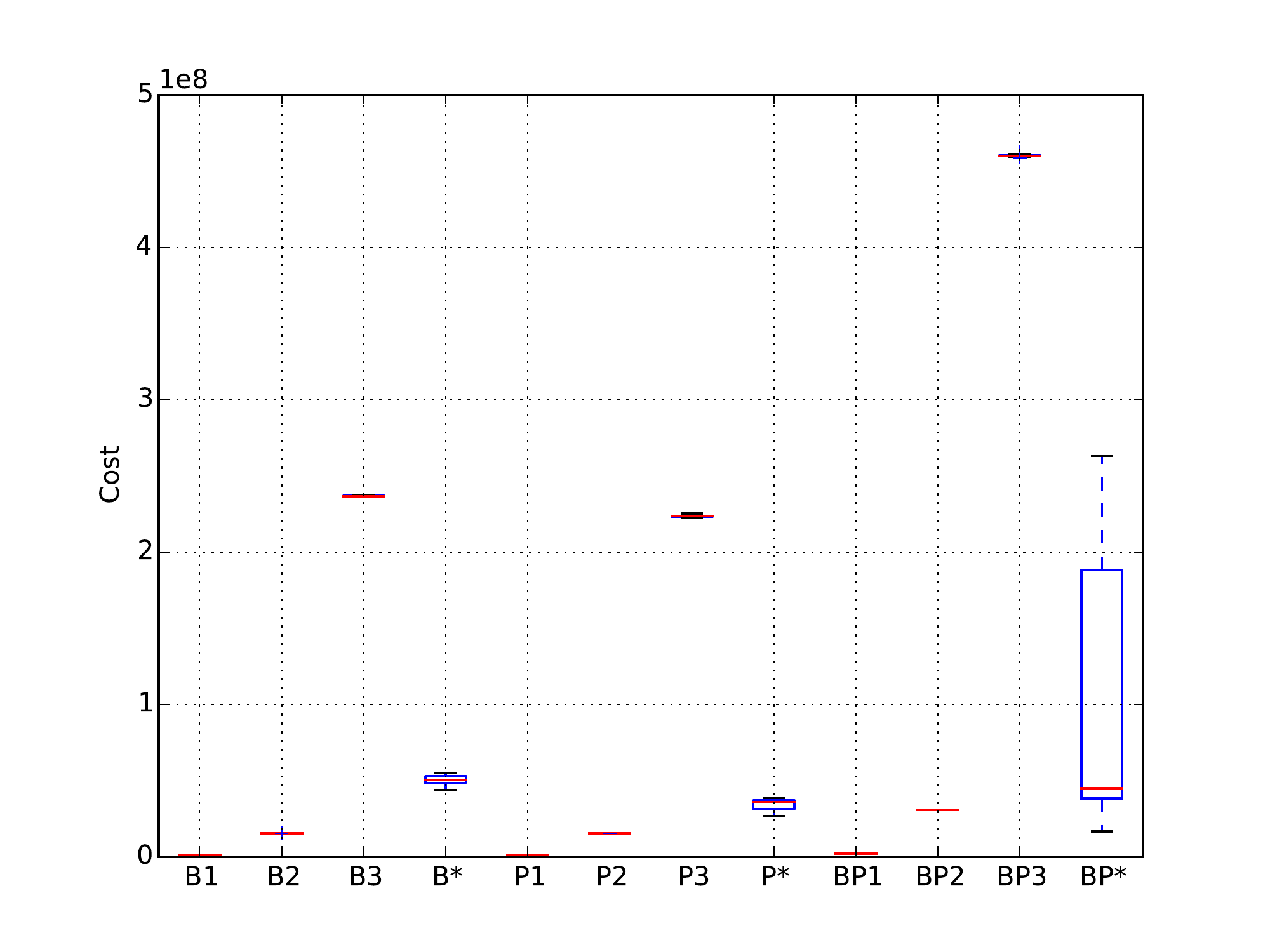}
   \caption{Overall time cost on the test set of the microscopic database, over 10 replications. 
   B1, B2, B3: SLF with LBP and $L$ set to 1, 2 
   and 3, respectively; B*: AMLF with LBP feature set; the same notation is used for P1, P2, P3, P*, for the LPQ feature
   set, and BP1, BP2, BP3, BP* for both LBP and LPQ combined. }
   \label{fig:cost_micro}
\end{figure}

Besides that, the main advantage of the proposed AMLF becomes evident when the cost analysis is 
carried out, the results of which are depicted in Figure~\ref{fig:cost_micro}. In this case, SLF 
with LBP and LPQ combined and $L=3$ was on average about 10 times slower than AMLF with the same 
feature set. With LBP and LPQ individually, AMLF was generally almost 5 times faster than SLF. Note that the
complexity can vary with the partitioning of the data set, especially in the experiment where
the two feature sets were combined. Even with this variability, AMLF is at least twice as faster
than SLF, but it can be also 16 times faster.

\begin{figure}[H]
   \centering
   \includegraphics[width=8cm]{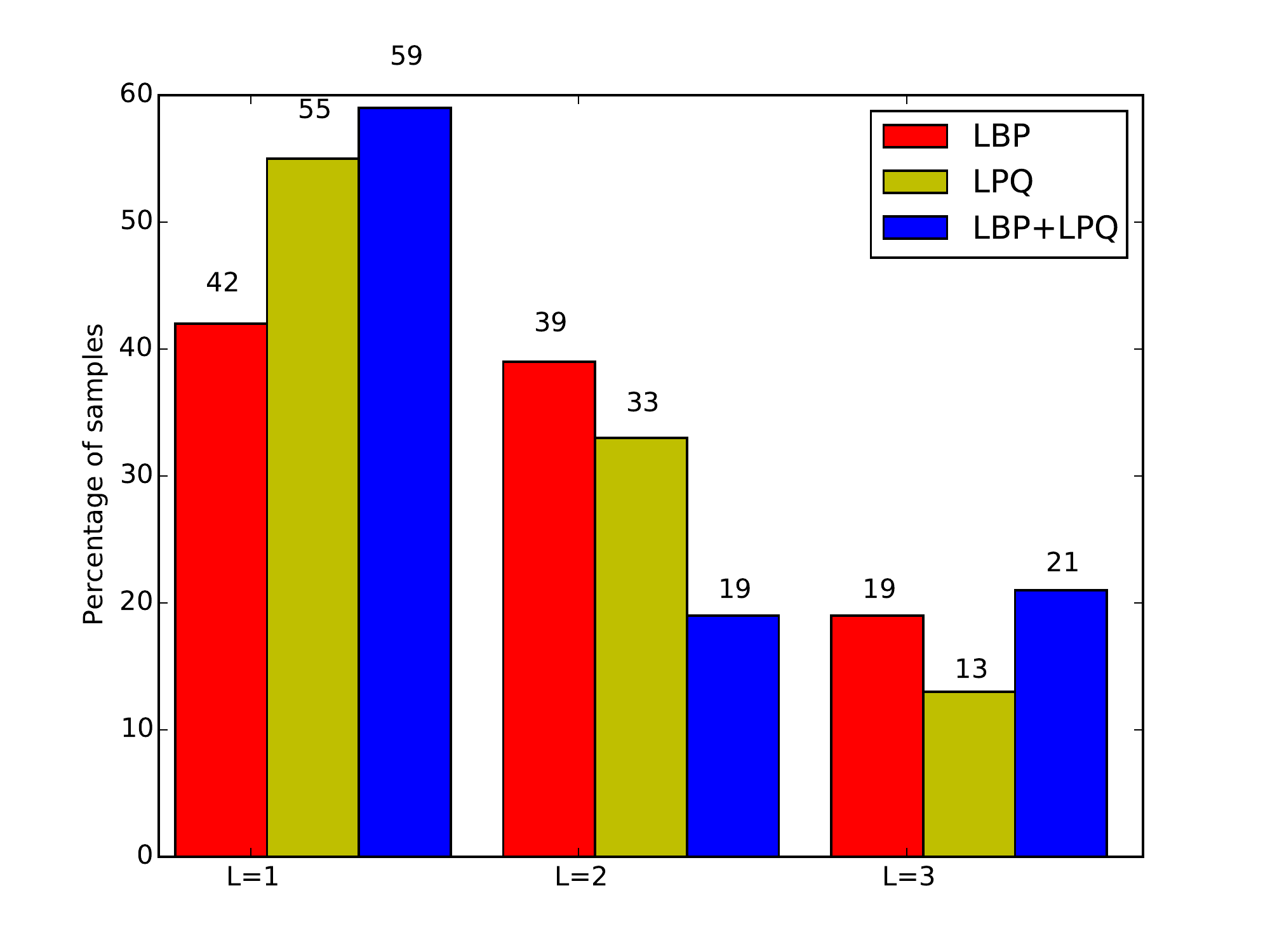}
   \caption{Average percentage of samples recognized at each level of AMLF with LBP, LPQ, and 
   LBP and LPQ combined, respectively, on the microscopic dataset.}
   \label{fig:bar_perc_micro}
\end{figure}

In order to complement this evaluation, in Figure~\ref{fig:bar_perc_micro} we present the average number 
of samples recognized in each level of AMLF (in which level the system stopped). We observe that when only 
one feature set is used, generally about for half of the samples it stops at level 1. Then, for about 33-39\% 
of the samples it stops at level 2. And only about 13-16\% of the samples were classified by level 3. With both 
LBP and LPQ feature sets combined, though, we observe that more samples are recognized at level 1, which is
most given to the stronger classification scheme that is made by the combination of the feature sets.

\subsubsection{Results on the Macroscopic dataset}
\begin{figure}[H]
   \centering
   \includegraphics[width=9cm]{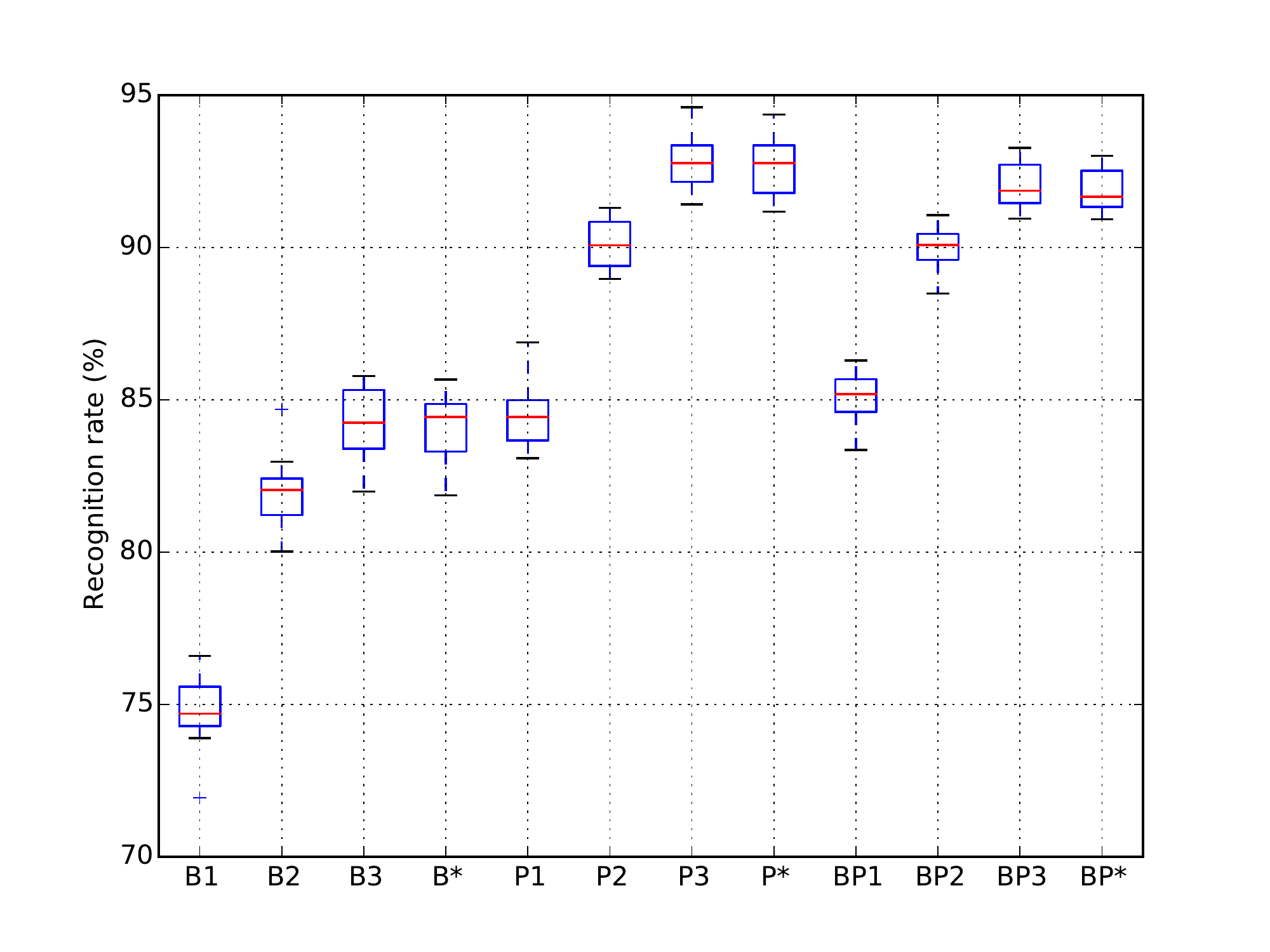}
   \caption{Recognition rates on the test set of the macroscopic database, over 10 replications. 
   B1, B2, B3: SLF with LBP and $L$ set to 1, 2 and 
   3, respectively; B*: AMLF with LBP feature set; the same notation is used for P1, P2, P3, P*, for the LPQ feature
   set, and BP1, BP2, BP3, BP* for both LBP and LPQ combined.}
   \label{fig:rec_rates_macro}
\end{figure}

The recognition rates of the different systems, evaluated on the macroscopic dataset, are presented in 
Figure~\ref{fig:rec_rates_macro}. Similarly to the results on the other dataset, SLF achieves slightly
higher recognition rates. The best recognition rates, of about 92.81\%, are achieved with SLF with
LPQ features only and $L$ set to 3. With LBP features only, and LBP and LPQ combined, the accuracy
was of 84.14\% and 92.00\%, respectively. The recognition rates of AMLF for LBP only, LPQ only, and
LBP and LPQ combined, were of 83.97\%, 92.63\%, 91.86\%. It is worth mentioning that there is
a smaller difference between the results of SLF and AMLF with this dataset, making it more evident that 
the latter might present a performance that is similar to a costly version of the former.

\begin{figure}[H]
   \centering
   \includegraphics[width=9cm]{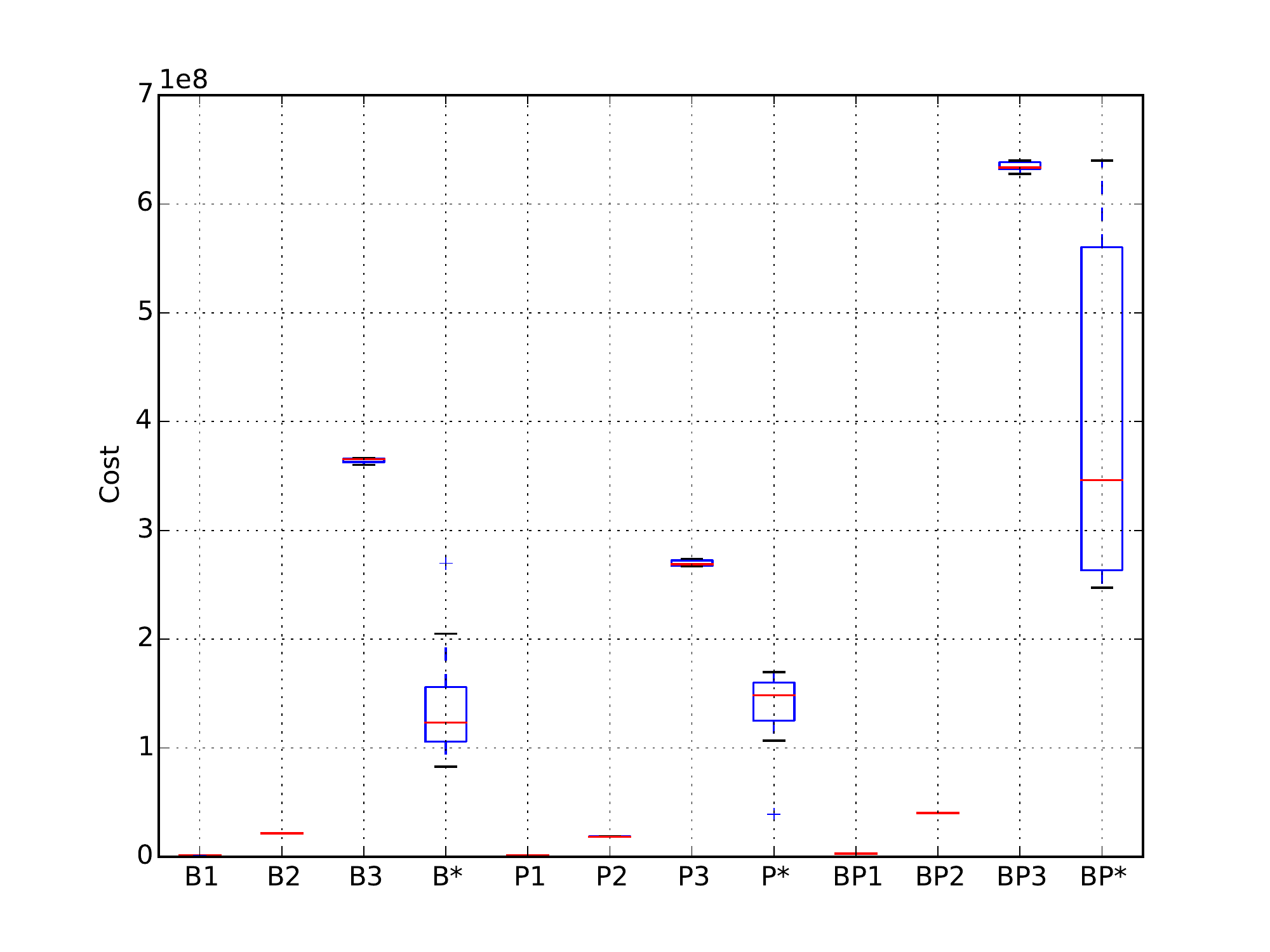}
   \caption{Overall time cost on the test set of the macroscopic database, over 10 replications. 
   B1, B2, B3: SLF with LBP and $L$ set to 1, 2 
   and 3, respectively; B*: AMLF with LBP feature set; the same notation is used for P1, P2, P3, P*, for the LPQ feature
   set, and BP1, BP2, BP3, BP* for both LBP and LPQ combined. }
   \label{fig:cost_macro}
\end{figure}

The cost analysis for this dataset is presented in Figure~\ref{fig:cost_macro}. In this case,
though, we observe that AMLF is not as faster as it can be in the microscopic dataset. 
With the best feature set, i.e. LPQ, AMLF is on average twice as faster than SLF with $L=3$.
With LBP, it is on average 3 times faster. And with LBP and LPQ combined, AMLF is on average
twice as faster, but depending on the partitioning of the dataset, it can be about 3 times faster
or even have the same cost.

\begin{figure}[H]
   \centering
   \includegraphics[width=8cm]{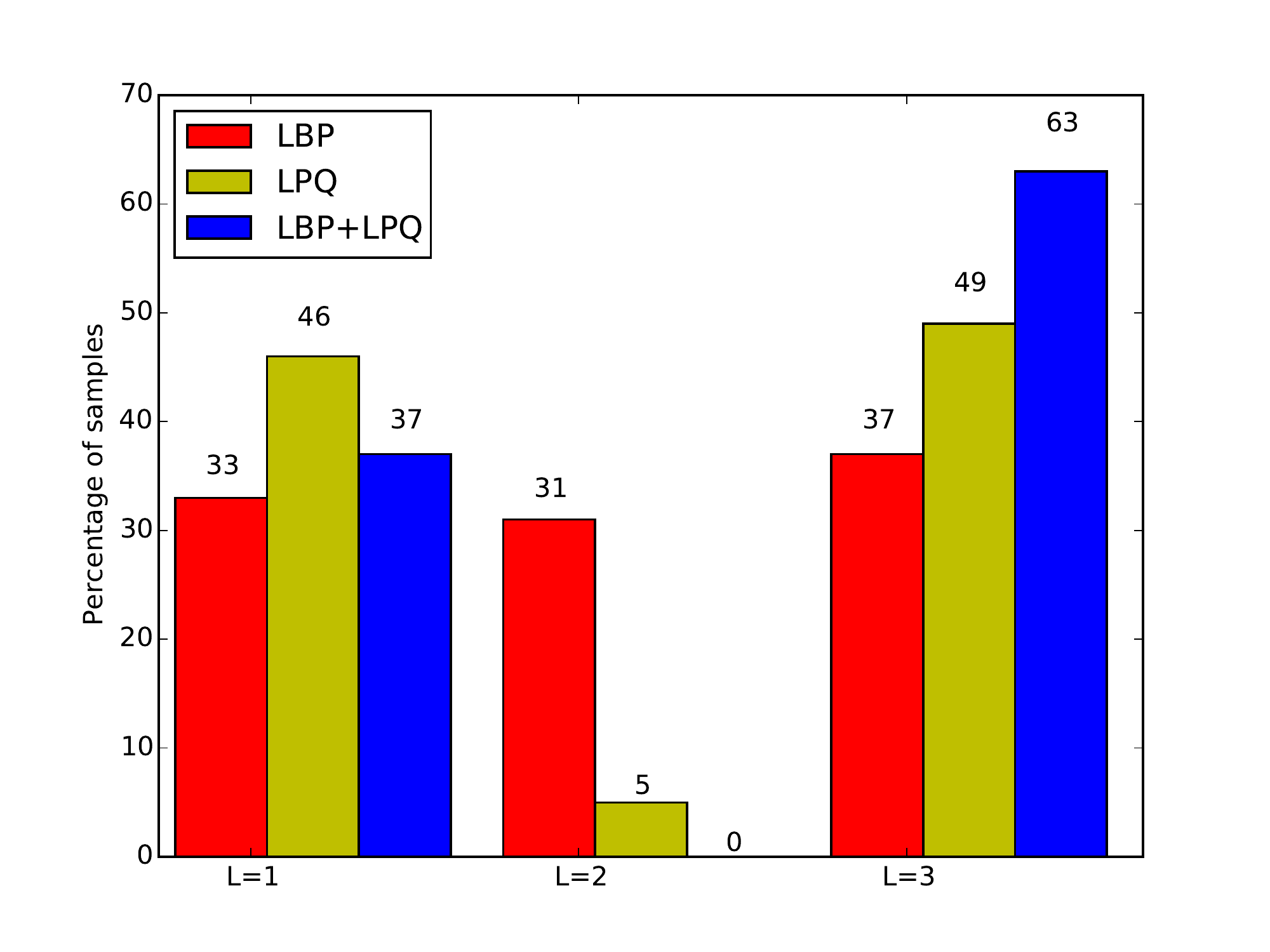}
   \caption{Average percentage of samples recognized at each level of AMLF with LBP, LPQ, and 
   LBP and LPQ combined, respectively, on the macroscopic dataset.}
   \label{fig:bar_perc_macro}
\end{figure}

The reason for the smaller difference in cost in this set can be easily visualized in Figure~\ref{fig:bar_perc_macro}.
In this case, we can observe that a similar amount of samples is recognized at level 1, i.e. from 33 to 46\%.
Nevertheless, much less samples are recognized at level 1, i.e. from 0 to 31\%, and more samples at level 2, i.e.
from 37 to 63\%. With more samples being recognized at level 2, the more costly level, the cost of AMLF tends 
to get closer to the cost of SLF with $L=3$.

\section{Feature Extraction Cost Reduction}
\label{sec:feature_extraction_cost_reduction}
In this section, the focus is shifted to cost reduction at feature extraction level. The main
idea to reduce this cost is pretty straight-forward, i.e. the image resolution can be reduced
at a given scale $S$, where $0 \le S \le 1$, and the feature extraction cost is reduced in
the same proportion.

To make this idea clearer, in Section~\ref{sec:feature_extraction_cost}, we discuss in greater 
detail how resolution reduction affects the cost of feature extraction. Next, in 
Section~\ref{sec:feature_extraction_experiments}, we present the experiments that have been 
conducted to evaluate the impact on the accuracy of SLF.

\subsection{Feature Extraction Cost Definition}
\label{sec:feature_extraction_cost}
As we mentioned, feature extraction cost reduction can be directly measured by means of
the scale factor $S$ acting in the resolution of the image. In other words, the 
feature extraction phase can be simplified as extracting features for all $P$ pixels in the 
image, and the larger the value of $P$, the more costly is this phase. Thus, the cost of 
feature extraction can be defined as:
\begin{equation}
   \label{eq:cost_fe}
   Cost^F = P.
\end{equation}

And given that in this particular section we do not consider the cost of classification, the
overall cost $Cost''$ can be defined as:
\begin{equation}
   \label{eq:overall_cost_fe}
   Cost'' = \sum_{i=1}^{Ne}  Cost^F.
\end{equation}

Nonetheless, if the resolution of the original image, containing $P$ pixels, is reduced by a scale factor
$S$, where $0 \le S \le 1$, then $Cost^F$ can be defined as a function of $S$:
\begin{equation}
   \label{eq:cost_fe_s}
   Cost^F(S) = S \times P.
\end{equation}

Note that $S$ consists only of a multiplication factor directly affecting $Cost^F$, defined in 
Equation~\ref{eq:cost_fe}. For this reason, $Cost^F(S)$ could be simply simplified to:
\begin{equation}
   \label{eq:cost_fe_s_simplified}
   Cost^F(S) = S \times Cost^F.
\end{equation}

And the overall cost can be simply defined as:
\begin{equation}
   Cost''(S) = S \times Cost'',
\end{equation}
\noindent considering $Cost''$ defined in Equation~\ref{eq:overall_cost_fe}.

\subsection{Experiments}
\label{sec:feature_extraction_experiments}
In this section we present the experiments to validate the impact of feature extraction cost
reduction in the accuracy of different implementations of SLF, with $L$ ranging from 1 to 3,
and with LBP, LPQ, and LBP and LPQ combined. The protocol used herein is the same as the one defined
in Section~\ref{sec:experiments_protocol}.

\begin{figure}[H]
   \centering
   \includegraphics[width=9cm]{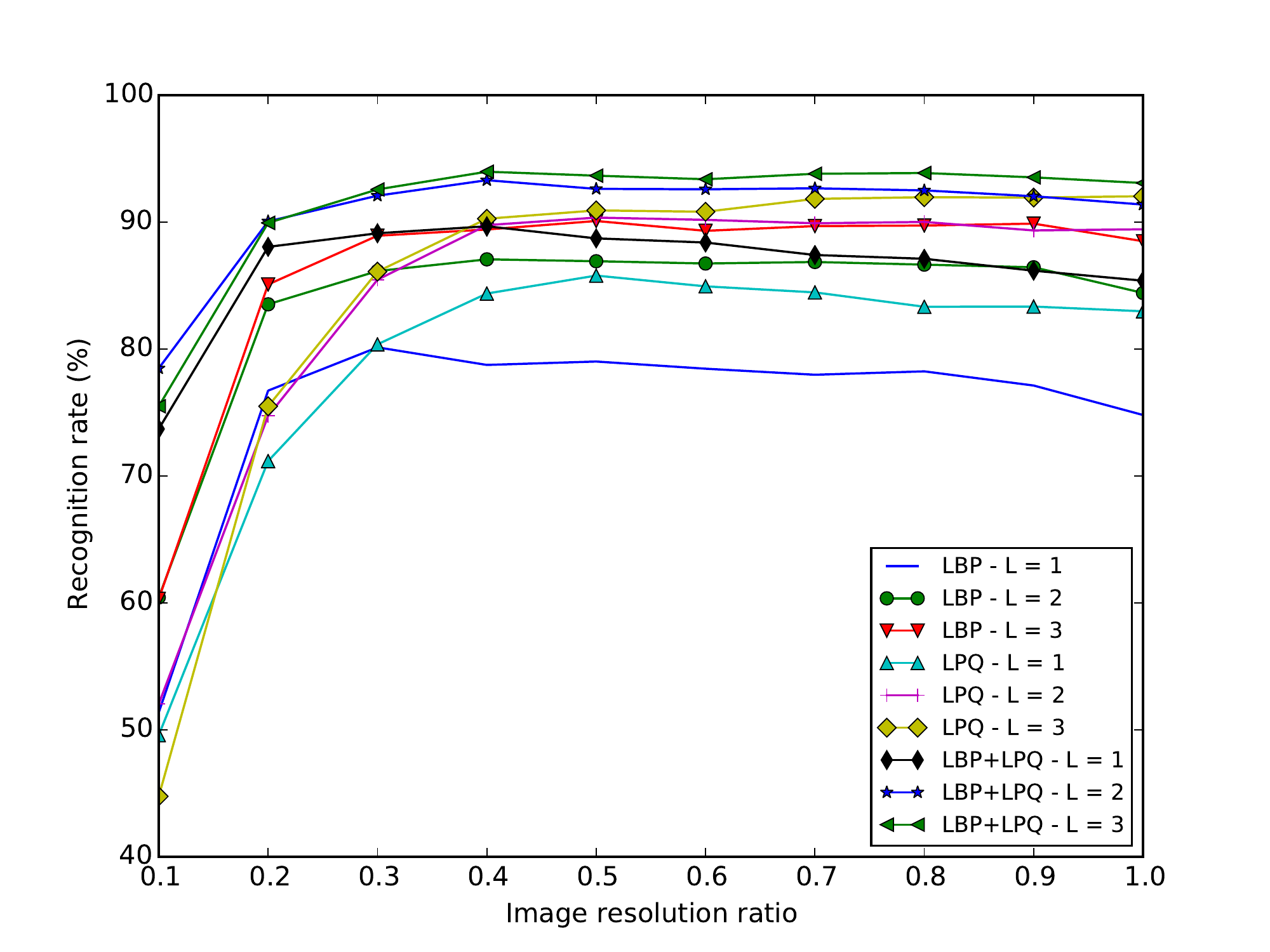}
   \caption{Average recognition rates with resolution scale $S$ from 0.1 to 
   1.0, in the microscopic dataset.}
   \label{fig:low_res_micro}
\end{figure}

In Figure~\ref{fig:low_res_micro} we provide the results on the microscopic dataset. We can observe
that the feature extraction cost can be significantly reduced while maintaining or even surpassing
the original recognition rates. Considering both LBP and LPQ combined, with $L=3$, the cost
can be reduced to 40\% ($S=0.4$) and better recognition rates are achieved, i.e. 93.97\% compared
with the 93.08\% for $S=1.0$. For smaller values of $L$, the cost can be reduced to even smaller
levels with no loss of performance. With $L$ set to 1, $S=0.2$ is about 2.67 percentage points 
better than $S=1.0$. And with $L$ set to 2, $S=0.3$ is about 0.7 percentage points better than
$S=1.0$. A similar scenario can be observed with LBP, where the best recognition rates are
achieved with $L$ set to 3 and $S=0.5$, reaching recognition rates of about 90.09\%, compared with
the 88.49\% achieved with $S=1.0$. With LPQ, however, gains are only observed with $L$ set to either
1 or 2. With $L$ set to 3, $S$ can only be set to about 0.7 to keep the accuracy loss at
minimum level.

\begin{figure}[H]
   \centering
   \includegraphics[width=9cm]{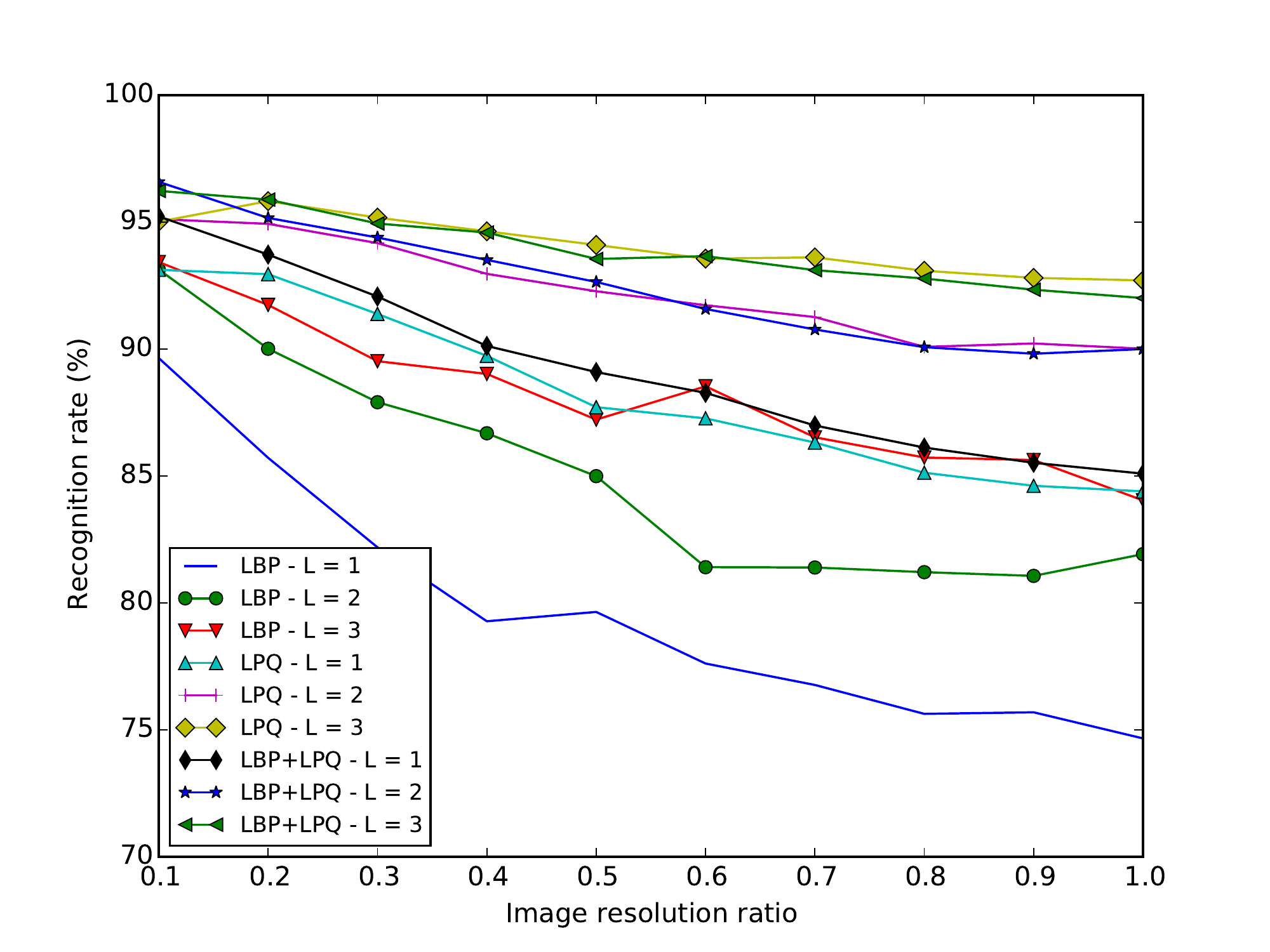}
   \caption{Average recognition rates with resolution scale $S$ from 0.1 to 
   1.0, in the macroscopic dataset.}
   \label{fig:low_res_macro}
\end{figure}

The corresponding evaluations on the macroscopic dataset are presented in Figure~\ref{fig:low_res_macro}.
In this case, reducing the cost of feature extraction generally results also on a positive
impact on the recognition rates. Considering the combination of the both feature sets, the
best recognition rates are achieved with $L=2$ and $S=0.1$, with recognition rates of about
96.58\%. Compared with the best results with the original image, i.e. $L=3$ and $S=1.0$, this
represents a gain of about 4.58 percentage points. In this case, for each configuration of $L$,
$S=0.1$ always results in the best recognition rates, being the impact more significant on
smaller values of $L$. With the feature sets used individually we observe a similar behavior.
Considering LBP with $L=3$, $S=0.1$ represents a gain of 9.93 percentage points compared with 
$S=1.0$. And  considering LPQ with $L=3$, $S=0.2$ results in a gain of 3.12 percentage points
compared with $S=1.0$.

\section{Global Cost Reduction}
\label{sec:global_cost_reduction}
After the presentation of methods that can be used successfully to reduce costs at either 
classification or feature extraction levels, in this section we evaluate how cost can be 
reduced for both levels at the same time, i.e. globally. That means, basically,
the application of AMLF on images reduced by the scale factor $S$, and the evaluation of
the impact on the recognition rates and on the cost.

To achieve this goal, we first describe a way to compute global cost, followed by the 
experimental evaluation.

\subsection{Global Cost Definition}
\label{sec:global_cost}
The fundamental issue related to computing cost at global level is the balance between the terms that
represent the cost for feature extraction, i.e. $Cost^F$, and the cost for classification, i.e. $Cost^C$,
in the $Cost$ defined in Equation~\ref{eq:cost}. One solution that we propose is to take into
account the dimension of the vectors involved in each phase, and the number of times a
basic operation is applied on them. In this case, the basic operation is a multiplication
involved in the dot product of two vectors. Let us first explain how this idea can be employed
to compute $Cost^F$ and $Cost^C$, respectively, and then to compute the global costs for both
SLF and AMLF.

By extending the ideas presented in Section~\ref{sec:feature_extraction_cost}, and considering
that the feature extraction for all $P$ pixels involves applying a filter on the neighbourhood 
window of size $W$, the cost of
feature extraction presented in Equation~\ref{eq:cost_fe} can be extended in the following 
way:
\begin{equation}
   \label{eq:cost_fe_dec}
   Cost^F(S) = S \times (P \times W).
\end{equation}

Similarly, for classification, we can add the $D$ term to represent the dimension of the
vector inputed to the classifier, and define $Cost^C$ as:
\begin{equation}
   \label{eq:cost_c_dec}
   Cost^C = \Gamma \times D,
\end{equation}
\noindent where $\Gamma$ represents the cost of the base classifier, for instance the number of
support vector in an SVM classifier.

Considering that the classification can take into account multiple classifiers, and the number 
of classifiers and the cost of the base classifier can be a function of $L$, we can extend 
Equation~\ref{eq:cost_c_dec} to:
\begin{equation}
   \label{eq:cost_c_dec_l}
   Cost^C(L) = f(L) \times (\Gamma(L) \times D).
\end{equation}

As a result, the global cost for SLF, considering the scale factor $S$ and
number of classifiers as a function of $L$, can be defined as:
\begin{equation}
   Cost_{SLF}(L, S) = Ne \times (Cost^F(S) + Cost^C(L))
\end{equation}

Similarly, considering also the scale factor $S$ and the maximum level as $L_{max}$, the global
cost for AMLF can be defined as:
\begin{equation}
   Cost_{AMLF}(S, L_{max}) = \sum_{L=1}^{L_{max}} [ Ne_{L} \times \sum_{l=1}^{L} [Cost^F(S) + Cost^C(L)] ].
\end{equation}

\subsection{Experiments}
\label{sec:experiments_static}
In this section, we describe the experiments conducted to evaluate global cost reduction, using
the same experimental described in Section~\ref{sec:experiments_protocol}. Basically, we compare
the results of AMLF and SLF with $S=1.0$, with the best value of $S$ found in the results
presented in Section~\ref{sec:feature_extraction_experiments}, i.e. $S=0.4$ for microscopic
and $S=0.1$ for macroscopic.

\begin{figure}[H]
   \centering
   \includegraphics[width=9cm]{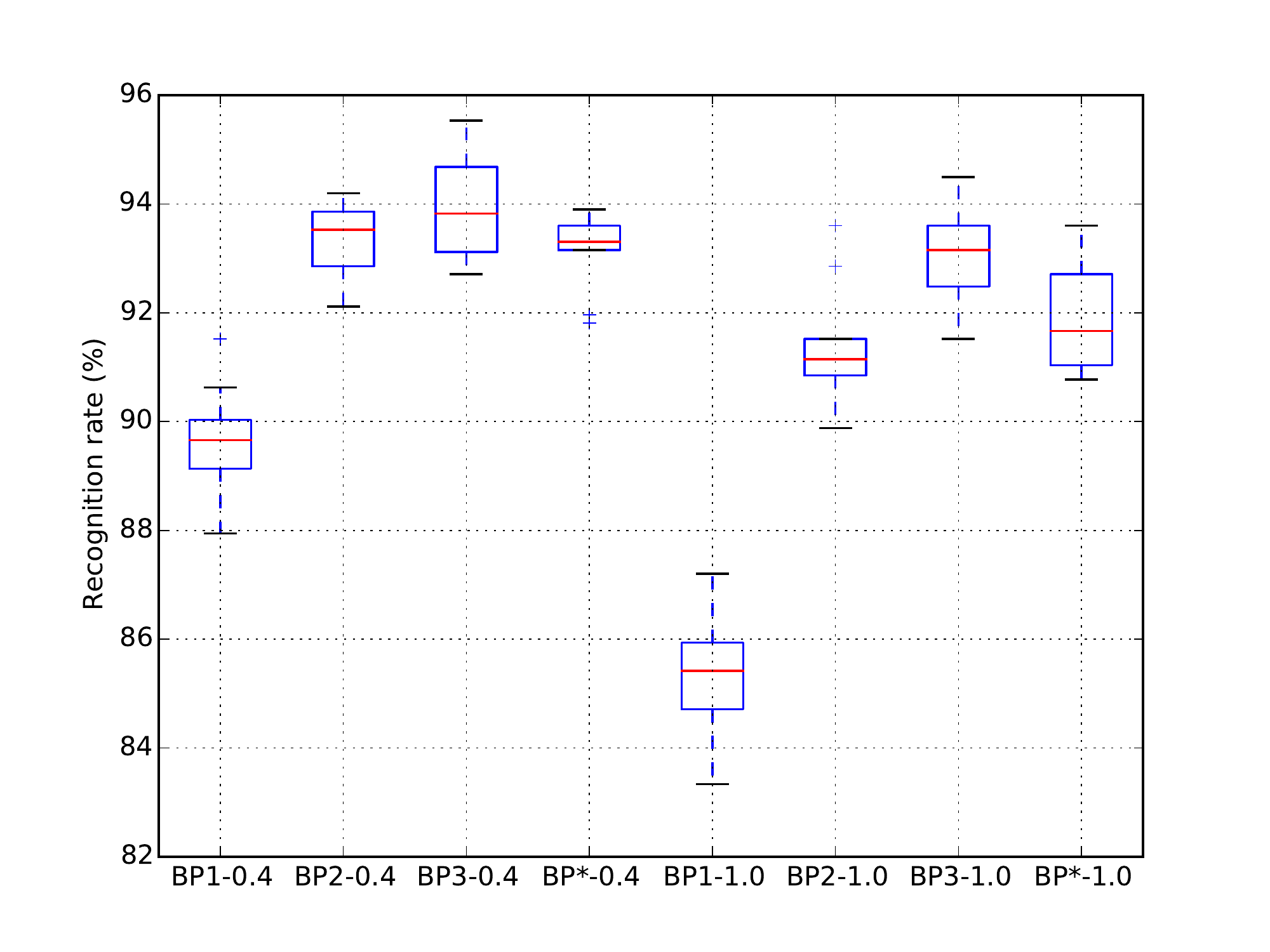}
   \caption{Recognition rates on the test set of the microscopic database, over 10 replications,
   for lower resolution system with the best results. 
   BP1, BP2, BP3: SLF with LBP and LPQ feature sets combined, with $L$ set to 1, 2 and 
   3, respectively; BP*: AMLF with LBP and LPQ feature sets combined.}
   \label{fig:rr_global_micro}
\end{figure}

\begin{figure}[H]
   \centering
   \includegraphics[width=9cm]{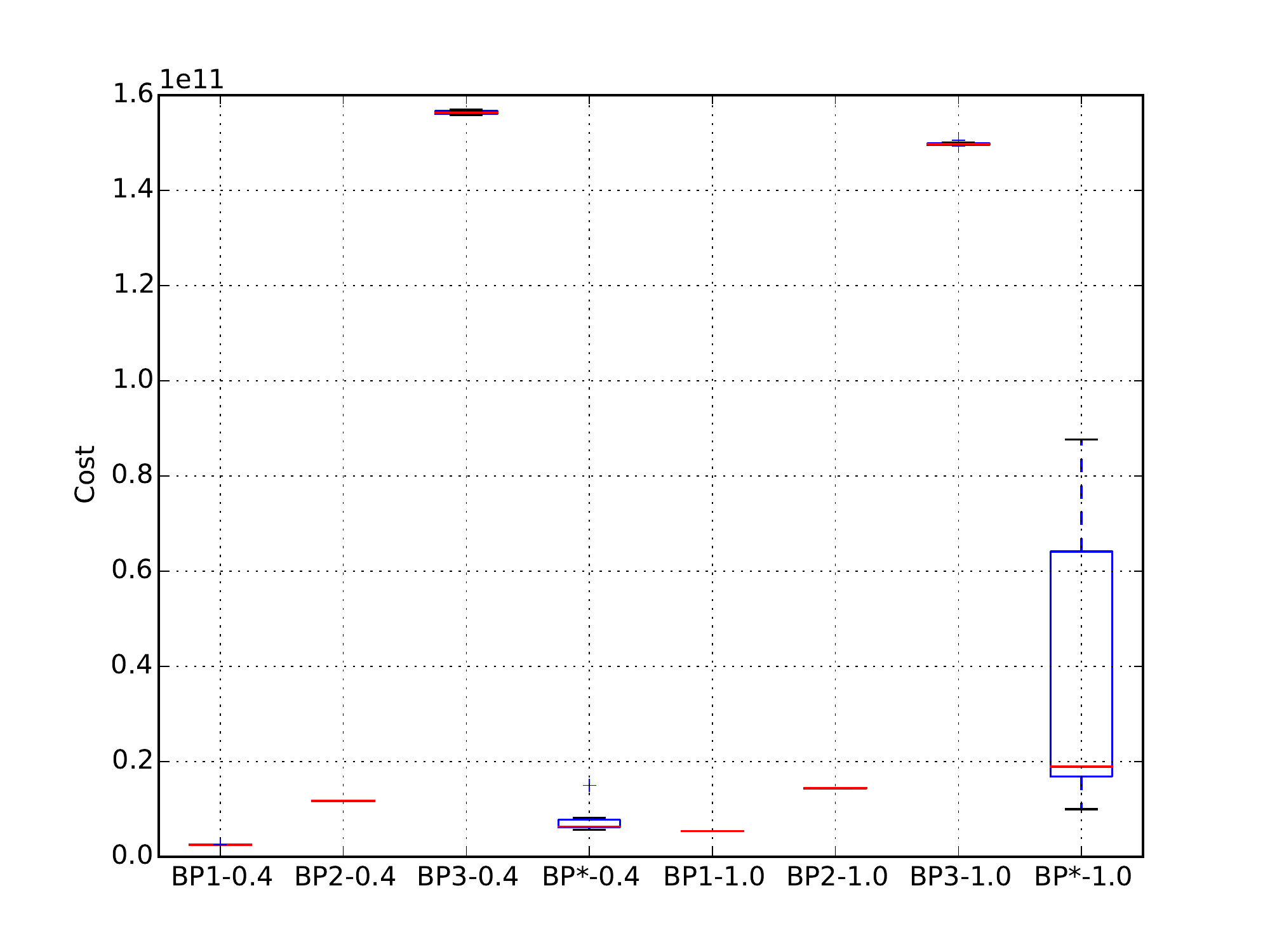}
   \caption{Overall time cost on the test set of the microscopic database, over 10 replications, 
   for lower resolution system with the best results. 
   BP1, BP2, BP3: SLF with LBP and LPQ feature sets combined, with $L$ set to 1, 2 and 
   3, respectively; BP*: AMLF with LBP and LPQ feature sets combined.}
   \label{fig:cost_global_micro}
\end{figure}

For the microscopic database, we present the recognition rates in Figure~\ref{fig:rr_global_micro} and
the costs in Figure~\ref{fig:cost_global_micro}. In terms of the recognition rates achieved by
AMLF, we observe an increase from 91.90\% to 93.17\%. Furthermore, the gap between AMLF and
SLF with $L=3$ decreases from 2.18 to 0.8 percentage points. In terms of cost, two aspects
are worth mentioning. The first is that the average global cost of AMLF is reduced to about
1/3 of the cost of the original system, i.e. $S=1.0$. Moreover, it is also interesting that
the costs of the different experiment replications present a much smaller standard deviation,
and the maximum cost is considerably smaller (about 1/4) with $S=0.4$. Finally, if we compare
the cost of AMLF with $S=0.4$ against SLF with $L=3$ and $S=1.0$, i.e. a system with global cost
reduction against a system with no reduction at all, the former presents about only 1/20 of 
the cost of the latter with better recognition rates.

\begin{figure}[H]
   \centering
   \includegraphics[width=9cm]{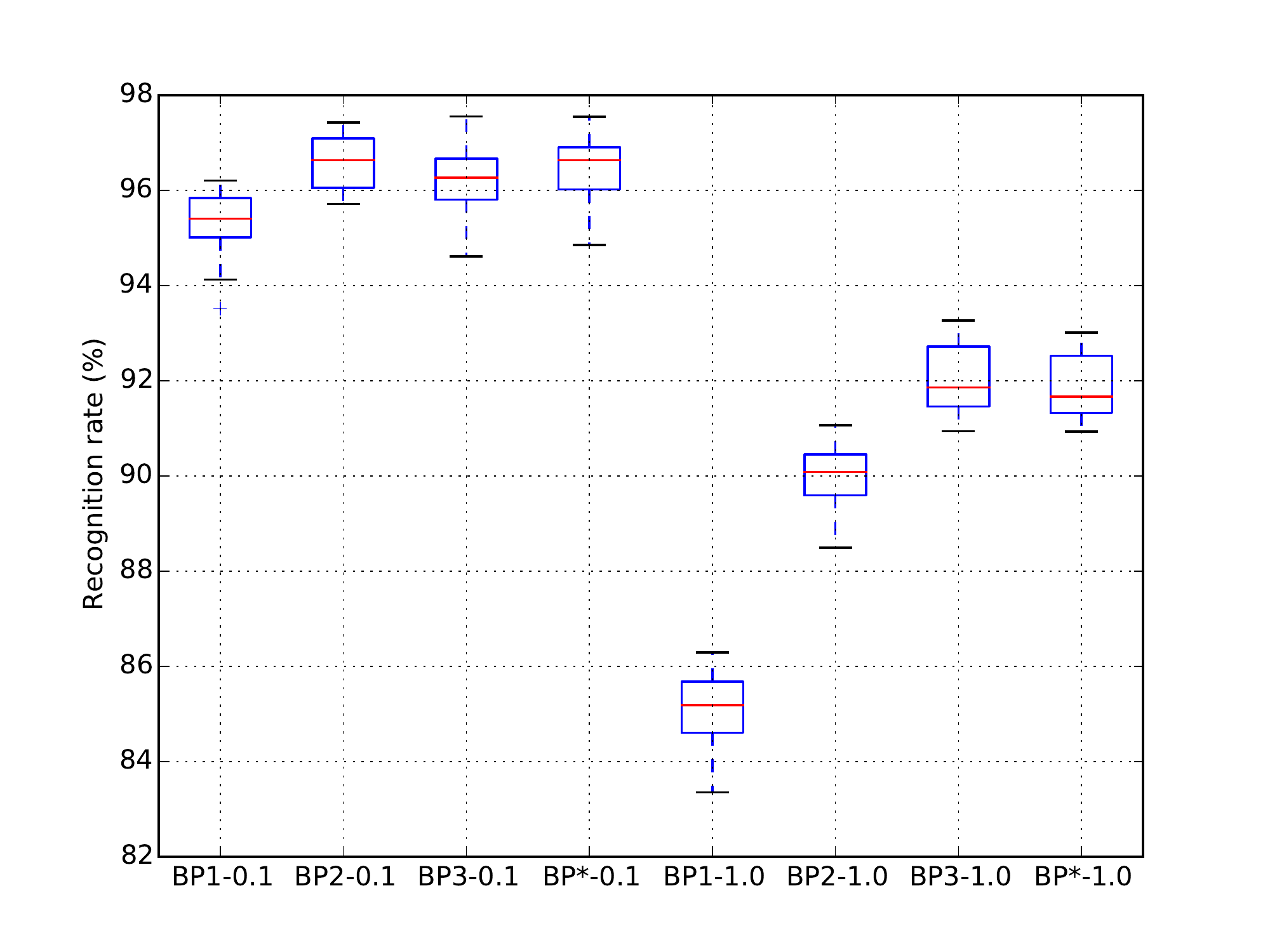}
   \caption{Recognition rates on the test set of the macroscopic database, over 10 replications,
   for lower resolution system with the best results. 
   BP1, BP2, BP3: SLF with LBP and LPQ feature sets combined, with $L$ set to 1, 2 and 
   3, respectively; BP*: AMLF with LBP and LPQ feature sets combined.}
   \label{fig:rr_global_macro}
\end{figure}

\begin{figure}[H]
   \centering
   \includegraphics[width=9cm]{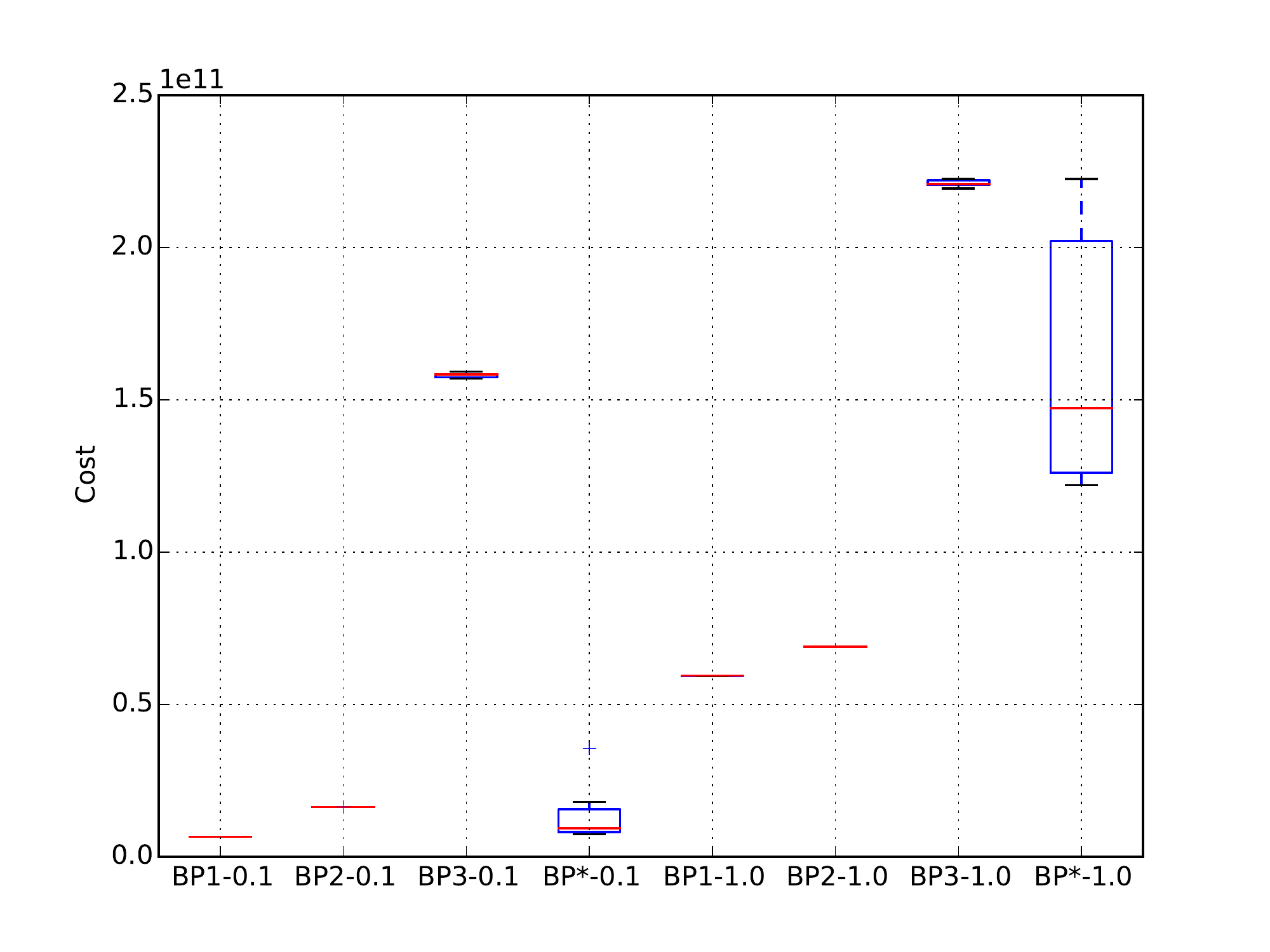}
   \caption{Overall time cost on the test set of the macroscopic database, over 10 replications, 
   for lower resolution system with the best results. 
   BP1, BP2, BP3: SLF with LBP and LPQ feature sets combined, with $L$ set to 1, 2 and 
   3, respectively; BP*: AMLF with LBP and LPQ feature sets combined.}
   \label{fig:cost_global_macro}
\end{figure}

The corresponding recognition rates and costs computed on the macroscopic dataset
are presented in Figure~\ref{fig:rr_global_macro} and Figure~\ref{fig:cost_global_macro}.
In this case, the impact of global cost reduction is considerably more visible in both 
aspects. A very significant increase in accuracy is observed with $S=0.1$. In this case, 
AMLF presents recognition rates of about 96.48\%, against 91.86\% reached with $S=1.0$. 
In terms of cost, with $S=0.1$, the average cost of the AMLF can be reduced to about 1/15 
of the average cost of AMLF with $S=1.0$. Again, comparing the system with global cost
reduction with the system with no reduction at all, the reduction in cost is of about 22
times with much better recognition rates.

\begin{figure}[H]
   \centering
   \includegraphics[width=8.5cm]{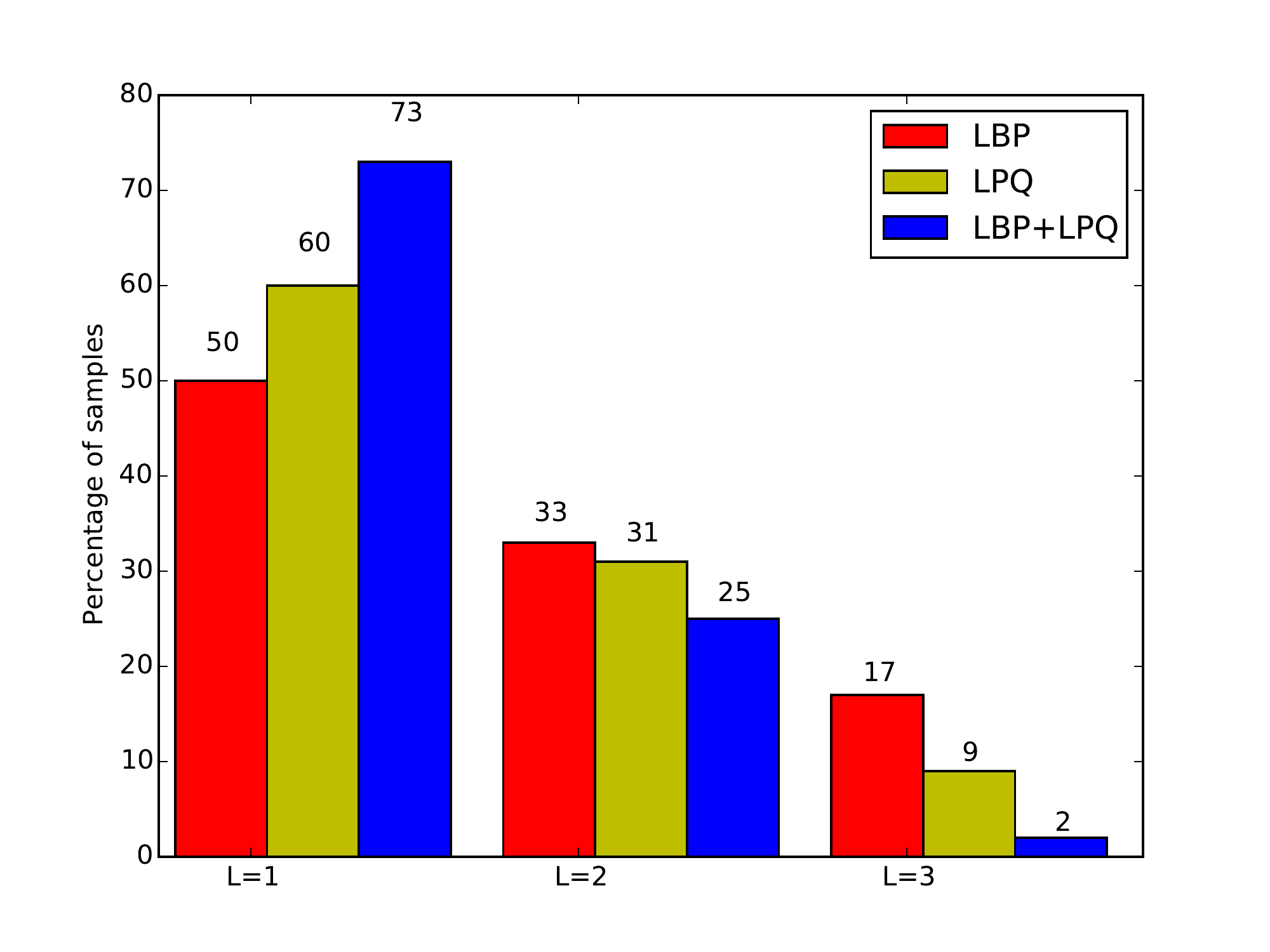}
   \caption{Average percentage of samples recognized at each level of the best configuration
   of AMLF with LBP, LPQ, and 
   LBP and LPQ combined, respectively, on the microscopic dataset.}
   \label{fig:bar_perc_micro_lowerres}
\end{figure}

\begin{figure}[H]
   \centering
   \includegraphics[width=8.5cm]{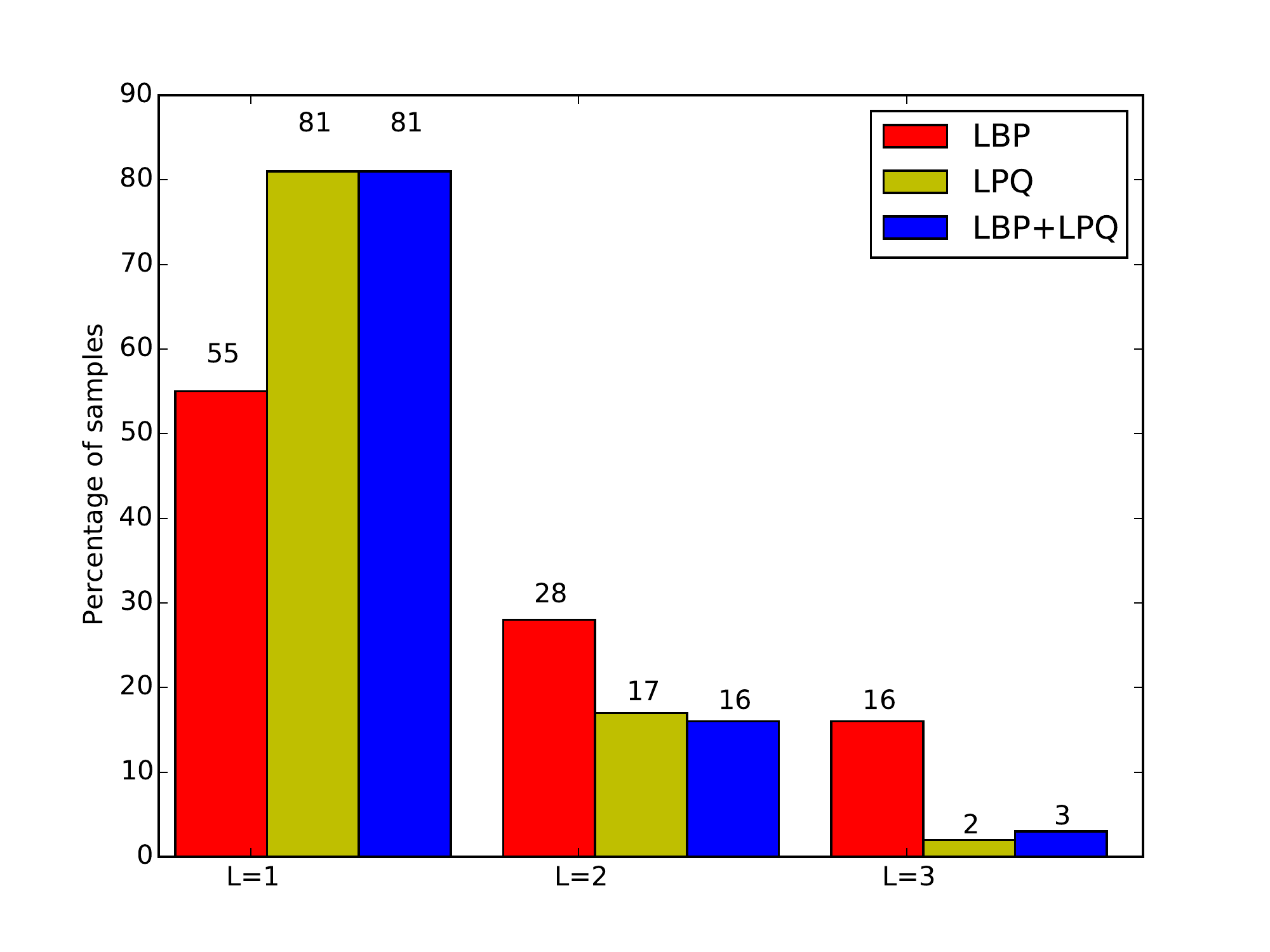}
   \caption{Average percentage of samples recognized at each level of the best configuration
   of AMLF with LBP, LPQ, and 
   LBP and LPQ combined, respectively, on the macroscopic dataset.}
   \label{fig:bar_perc_macro_lowerres}
\end{figure}

To conclude these analyses, Figure~\ref{fig:bar_perc_micro_lowerres} and 
Figure~\ref{fig:bar_perc_macro_lowerres} present the average percentage of samples recognized 
at each layer in AMLF, for microscopic and macroscopic respectively. Compared with
Figure~\ref{fig:rec_rates_micro} and Figure~\ref{fig:rec_rates_macro}, we observe that AMLF
tends to recognize more samples in the first layer with lower values of $S$, since the 
versions of SLF with lower values of $L$ are more accurate than those with higher values of $S$.
As a consequence, using SLF with $L=1$ for recognizing more samples and in a more accurate
way, has a direct impact in reducing the overall cost, while maintaining or even increasing 
the overall accuracy.

\section{Summary of the results}
In Table~\ref{tab:summary_results} we list the best results achieved in this work, in terms 
of both recognition accuracy and relative cost (using AMLF with reduced cost as reference), 
comparing against SLF (with no cost reduction) and the best methods from the literature\footnote{we present an estimate based on the resolution 
of the images and on the number of classifiers or classifications used by the method}.

\begin{table}[H]
  \centering
  \caption{\label{tab:summary_results} Results summary.}
  \begin{tabular}{|c|cc|cc|}
  \hline
               & \multicolumn{2}{c|}{\bf Microscopic dataset} & \multicolumn{2}{c|}{\bf Macroscopic dataset}\\
  \hline
  {\bf Method} & {\bf Accuracy (\%)} & {\bf Cost} & {\bf Accuracy (\%)} & {\bf Cost} \\
  \hline
  AMLF                   & 93.17 &  1.0 & 96.48 &  1.0 \\
  SLF                    & 93.08 & 20.0 & 92.81 & 22.0 \\
  Hafemann et al. \cite{Hafemann2014}    & 97.32 & 31.2 & 95.77 &  1.8 \\
  Paula Filho et al. \cite{PaulaFilho2014}  &     - &    - & 97.77 & 86.7 \\
  Kapp et al. \cite{Kapp2014}        & 95.68 & 60.0 & 88.90 & 22.0 \\
  \hline
  \end{tabular}
\end{table}

As observed, AMLF with global cost reduction can result in a system with about or less 
than 1/20 of the cost of SLF, but achieving better recognition rates in both datasets. 
Comparing with other methods that achieved better results on the microscopic dataset, 
we see that the methods presented in \cite{Hafemann2014} and 
\cite{Kapp2014} reach recognition rates that are 4.15 and 2.51 percentage points better 
than AMLF, respectively, but with corresponding costs that are 31.2 and 60.0 times higher.
However, neither of those methods outperform AMLF in the macroscopic dataset. On that
database, the approach from \cite{PaulaFilho2014} achieves the best accuracy,
which is 1.33 percentage points better than that of AMLF, with a cost that is 86.7 higher.
It is worth mentioning, however, that the techniques used by these methods from the
literature could also be used with AMLF, which may likely improve its performance.

\section{Conclusion and future work}
\label{sec:conclusion}
In this paper we investigated ways to reduce costs of forest species recognition systems. We 
focused on local cost reduction of classification or feature extraction individually, and both 
combined, i.e. globally. 
%The results demonstrated that by making use of global cost reduction,
%the overall cost can be reduced to less than 1/20 with better recognition rates.

To reduce costs at classification level, we proposed an adaptive multi-level framework for forest 
species recognition, based on extending the static single-layer framework proposed in 
\cite{Cavalin2013}. This approach demonstrated to be able to achieve comparable accuracy 
to that of the most costly version of SLF, but with about 1/10 of the cost on the microscopic 
dataset, and 1/3 on macroscopic images.

For feature extraction cost reduction, the simple idea of reducing the resolution of the input 
image results in linear cost reduction, and in some cases, higher accuracy. With the microscopic
dataset, better recognition rates are achieved with only about 40\% of the original cost. And
on macroscopic images, much better accuracy is observed with the cost reduced to only 10\%.

A further evaluation in global cost reduction demonstrated that when AMLF is applied on images 
with lower resolution, 
the resulting accuracy can be equivalent or even better than that of SLF, but with the cost
reduced by more than 20 times.

As future work, many directions can be followed. One might be improving the proposed AMLF 
method, and the methods that are used for its set up. For instance, other approaches 
to define the rejection between sub-sequent layers could be investigated, for instance, 
class-based thresholds. At feature extraction level, we should also investigate the impact
of image resolution reduction with other feature sets to better understand the different
scenarios for which this method can be used. Another interesting direction would be the
extension of the investigation presented in this work to other texture recognition problems,
and other features sets and classifiers, especially Convolutional Neural Networks.

\bibliographystyle{abbrv}
\bibliography{myreferences}

\begin{thebibliography}{10}

\bibitem{ArunPriyaC2012}
{ArunPriya C.} and Antony~Selvadoss Thanamani.
\newblock A survey on species recognition system for plant classification.
\newblock {\em International Journal of Computer Technology \& Applications},
  3(3):1132--1136, 2012.

\bibitem{Bremananth2009}
R.~Bremananth, B.~Nithya B, and R.~Saipriya.
\newblock Wood species recognition system.
\newblock {\em International Journal of Electrical and Computing Engineering},
  4(5):315--321, 2009.

\bibitem{Cavalin2015}
Paulo~R. Cavalin, Marcelo~N. Kapp, and Luiz E.~S. Oliveira.
\newblock An adaptive multi-level framework for forest species recognition.
\newblock In {\em Brazilian Conference on Intelligent Systems}, Natal, Brazil,
  2015.

\bibitem{Cavalin2013}
Paulo~R. Cavalin, Jefferson Martins, Marcelo~N. Kapp, and Luiz~E. Oliveira.
\newblock A multiple feature vector framework for forest species recognition.
\newblock In {\em The 28th Symposium on Applied Computing}, pages 16--20,
  Coimbra, Portugal, 2013.

\bibitem{Gasson2010}
Peter Gasson, Regis Miller, Dov~J. Stekel, Frances Whinder, and Kasia
  {Ziemi\'nska}.
\newblock Wood identification of dalbergia nigra (cites appendix i) using
  quantitative wood anatomy, principal components analysis and {na\"ive} bayes
  classification.
\newblock {\em Annals of Botany}, (105):45--56, 2010.

\bibitem{Hafemann2014}
Luiz~G. Hafemann, Luiz~S. Oliveira, and Paulo Cavalin.
\newblock Forest species recognition using deep convolutional neural networks.
\newblock In {\em Proceedings of 22nd International Conference on Pattern
  Recognition (ICPR), 2014 22nd International Conference on Pattern Recognition
  (ICPR)}, pages 1103--1107, 2014.

\bibitem{Kapp2014}
M.~Kapp, R.~Bloot, P.~R. Cavalin, and L.~S. Oliveira.
\newblock Automatic forest species recognition based on multiple feature sets.
\newblock In {\em International Joint Conference ou Neural Networks}, pages
  1296--1303, 2012.

\bibitem{Khalid08}
M.~Khalid, E.~L.~Y. Lee, R.~Yusof, and M.~Nadaraj.
\newblock Design of an intelligent wood species recognition system.
\newblock {\em IJSSST}, 9(3), 2008.

\bibitem{Khalid2011}
Marzuki Khalid, Rubiyah Yusof, and Anis Salwa~Mohd Khairuddin.
\newblock Tropical wood species recognition system based on multi-feature
  extractors and classifiers.
\newblock In {\em 2nd International Conference on Instrumentation Control and
  Automation}, pages 6--11, 2011.

\bibitem{Last2002}
M.~Last, H.~Bunke, and A.~Kandel.
\newblock A feature-based serial approach to classifier combination.
\newblock {\em Pattern Analysis \& Applications}, 5(4):385--398, 2002.

\bibitem{Martins2015}
J.~Martins, L.~S. Oliveira, A.~S. Britto-Jr, and R.~Sabourin.
\newblock Forest species recognition based on dynamic classifier selection and
  dissimilarity feature vector representation.
\newblock {\em Machine Vision and Applications}, 26(2):279--293, 2015.

\bibitem{Martins2012b}
J.~Martins, L.~S. Oliveira, and R.~Sabourin.
\newblock Combining textural descriptors for forest species recognition.
\newblock In {\em 38th Annual Conference of the IEEE Industrial Electronics
  Society (IECON 2012)}, 2012.

\bibitem{Martins2013}
J.~Martins, L.~S. Oliveira, and R.~Sabourin.
\newblock A database for automatic classification of forest species.
\newblock {\em Machine Vision and Applications}, 24(3):567--578, 2013.

\bibitem{Nasirzadeh2010}
M.~Nasirzadeh, A.~A. Khazael, and M.~B Khalid.
\newblock Woods recognition system based on local binary pattern.
\newblock {\em Second International Conference on Computational Intelligence,
  Communication Systems and Networks}, pages 308--313, 2010.

\bibitem{PaulaFilho2010}
P.~L. {Paula Filho}, L.~S. Oliveira, A.~S. Britto, and R.~Sabourin.
\newblock Forest species recognition using color-based features.
\newblock In {\em Proceedings of the 20th International Conference on Pattern
  Recognition}, pages 4178--4181, Instanbul, Turkey, 2010.

\bibitem{PaulaFilho2014}
P.~L. {Paula Filho}, L.~S. Oliveira, S.~Nisgoski, and A.~S. {Britto Jr.}
\newblock Forest species recognition using macroscopic images.
\newblock {\em Machine Vision and Applications}, 25(4):1019--1031, 2014.

\bibitem{Tou07}
J.~Y. Tou, P.~Y. Lau, and Proceedings~of Y.~H.~Tay.
\newblock Computer vision-based wood recognition system.
\newblock In {\em Int. Workshop on Advanced Image Technology}, 2007.

\bibitem{Tou2008}
J.~Y. Tou, Y.~H. Tay, and P.~Y. Lau.
\newblock One-dimensional grey-level co-occurrence matrices for texture
  classification.
\newblock {\em International Symposium on Information Technology (ITSim 2008)},
  pages 1--6, 2008.

\bibitem{Tou2009}
Jing~Yi Tou, Yong~Haur Tay, and Phooi~Yee Lau.
\newblock A comparative study for texture classification techniques on wood
  recognition problem.
\newblock In {\em Proceeding of the Fifth International Conference on Natural
  Computation}, pages 8--12, 2009.

\bibitem{Yadav2015}
A.~R. Yadav, R.~S. Anand, M.~L. Dewal, and S.~Gupta.
\newblock Multiresolution local binary pattern variants based texture feature
  extraction techniques for efficient classification of microscopic images of
  hardwood species.
\newblock {\em Applied Soft Computing}, 32:101--112, 2015.

\bibitem{Yadav2013}
A.~R. Yadav, M.~L. Dewal, R.~S. Anand, and S.~Gupta.
\newblock Classification of hardwood species using ann classifier.
\newblock In {\em National Conference on Computer Vision, Pattern Recognition,
  Image Processing and Graphics}, pages 1--5, 2013.

\bibitem{Yusof2013}
R.~Yusof, M.~Khalid, and A.~S.~M. Khairuddin.
\newblock Fuzzy logic-based pre-classifier for tropical wood species
  recognition system.
\newblock {\em Machine Vision and Applications}, 24(8):1589--1604, 2013.

\bibitem{yusof2010}
Rubiyah Yusof, Nenny~Ruthfalydia Rosli, and Marzuki Khalid.
\newblock Using gabor filters as image multiplier for tropical wood species
  recognition system.
\newblock {\em 12th International Conference on Computer Modelling and
  Simulation}, pages 284--289, 2010.

\end{thebibliography}

\end{document}